\documentclass[10pt,twocolumn,letterpaper]{article}
\usepackage[pagenumbers]{cvpr}

\usepackage{xr-hyper}

\usepackage{cuted} %

\usepackage{currfile} %

\usepackage{caption} %

\usepackage{array}
\usepackage[above]{placeins}
\definecolor{cvprblue}{rgb}{0.21,0.49,0.74}
\usepackage[breaklinks,colorlinks,allcolors=cvprblue]{hyperref}
\def\confName{CVPR}
\def\confYear{2027}

\title{WorldPhy: Benchmarking Fine-Grained Physical Plausibility of World Models in Robotics}

\author{}
\hypersetup{pdfauthor={}}
\author{Ziyuan Chu$^{1}$\quad Mingzhi Zheng$^{2}$\quad Kaiyue Zhang$^{1}$\quad
        Jinan Zhang$^{1}$\quad Linchao Zhu$^{1,*}$\\
        $^{1}$Zhejiang University\qquad $^{2}$Nanjing University\qquad
        $^{*}$Corresponding author}
\hypersetup{pdftitle={\thetitle},pdfauthor={Ziyuan Chu, Mingzhi Zheng, Kaiyue Zhang, Jinan Zhang, Linchao Zhu}}

\makeatletter
\patchcmd{\@maketitle}{\vskip .375in}{\vskip .12in}{}{}
\patchcmd{\@maketitle}{\vspace*{24pt}}{\vspace*{12pt}}{}{}
\newcommand{\appendixcontents}{\section*{Appendix Contents}\@starttoc{toc}}
\newcommand{\appendixfloatlayout}{%
  \setlength{\@fptop}{0pt}%
  \setlength{\@dblfptop}{0pt}%
  \setlength{\@fpsep}{\textheight}%
  \setlength{\@dblfpsep}{\textheight}%
  \setcounter{topnumber}{1}%
  \setcounter{dbltopnumber}{1}%
  \AddToHook{cmd/@floatplacement/after}{%
    \ifx\@dbltoplist\@empty\else\suppressfloats[t]\fi}}
\makeatother

\begin{document}
\addtocontents{toc}{\protect\setcounter{tocdepth}{-10}}
\maketitle
\begingroup
\begin{strip}
  \captionsetup{type=figure}
  \centering
  \includegraphics[width=0.92\textwidth]{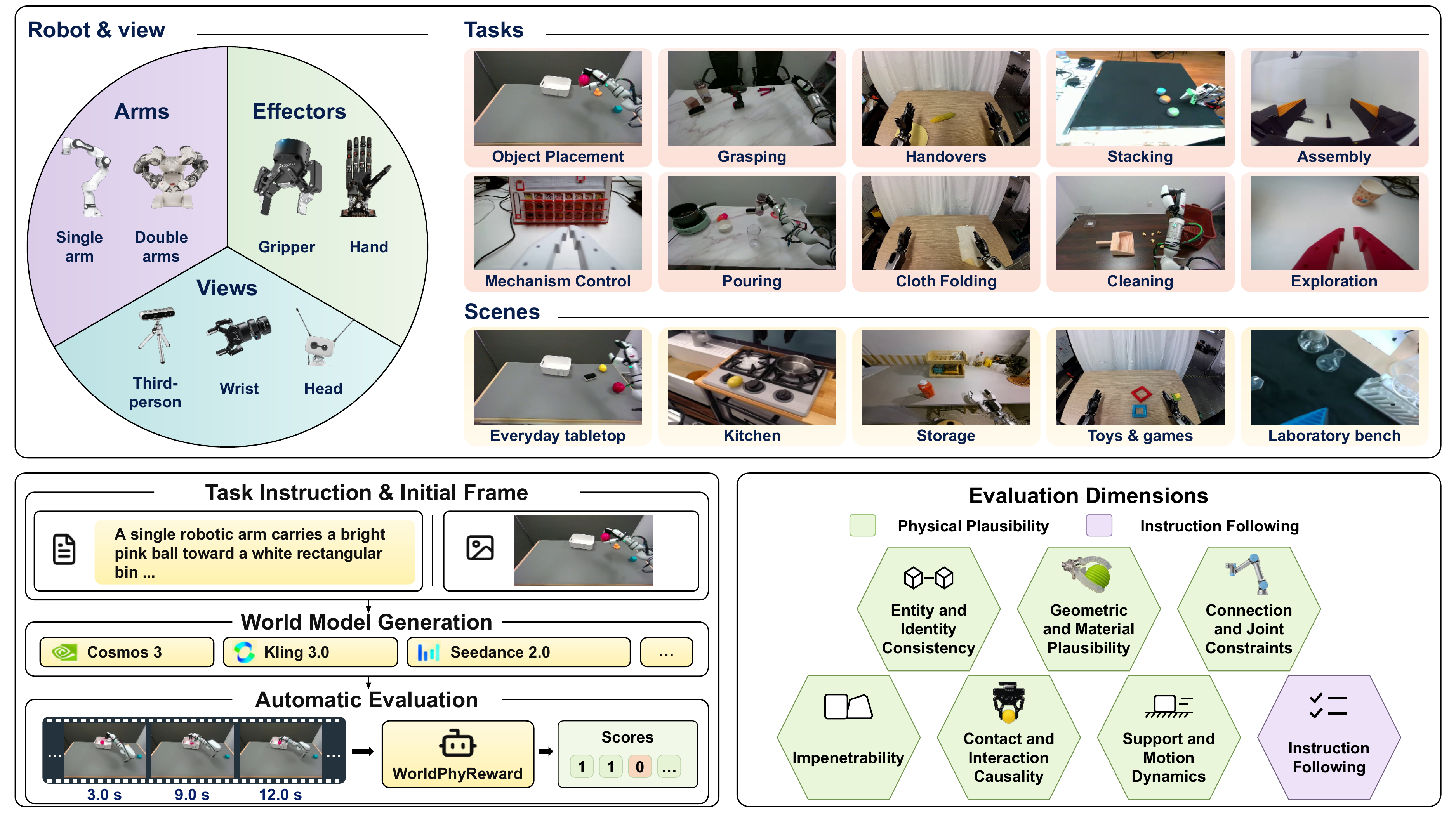}
  \caption{\textbf{Overview of WorldPhy.} Our benchmark evaluates world models in robotics across two arm configurations, three viewpoints, two end-effector types, ten task categories, and five scene categories. Each task provides an instruction and an initial frame to condition video generation. Our trained reward model, WorldPhyReward, then automatically evaluates the generated video, providing scores and supporting reasons over six fine-grained physical dimensions and an instruction-following dimension.}
  \label{fig:overview}
\end{strip}
\endgroup
\newcommand{\WorldPhyCodeAvailability}{Our code is available at \url{https://github.com/chuziyuan139/WorldPhy}.}
\begin{abstract}
World models are increasingly used in robotics to improve the performance of embodied models. Compared with general-purpose video generation, robotics places more stringent requirements on world models. Generated videos must satisfy fine-grained physical constraints on robot motion and object interactions. However, existing world model benchmarks can overlook robotics specific physical details and struggle to assess them accurately. To address this problem, we introduce WorldPhy, a world model benchmark for accurate, fine-grained evaluation of physical plausibility in robotics. Our benchmark contains 300 tasks, each consisting of an initial-frame image and a task instruction, covering diverse robotics scenes, task categories, viewpoints, and robot configurations. We further collect over 12K high-quality human annotations, and use them to train WorldPhyReward to accurately provide scores and reasons across six fine-grained physical dimensions and an instruction-following dimension. Comparisons with general VLMs and existing reward models show that WorldPhyReward better aligns with human ratings. We systematically evaluate 13 frontier open and closed-source world models. The results reveal that current world models struggle to consistently maintain fine-grained physical plausibility across different robotics scenarios.
\ifdefined\WorldPhyCodeAvailability
  \WorldPhyCodeAvailability
\else
  Our code, data, and model will be open-sourced.
\fi
\end{abstract}
\section{Introduction}
\label{sec:introduction}

World models predict future frames from observations, language instructions, or actions to support planning~\cite{agarwal2025cosmos}. In robotics, they improve embodied models by generating training data and predicting future frames for policy selection~\cite{jang2025dreamgen,yang2023learning,ebert2018visual}.

Robotics differs from general-purpose video generation and places more stringent requirements on world models. To be useful in robotics, generated videos must satisfy a range of fine-grained physical criteria, such as robot joint rotation limits, the falling speeds of objects, and gripper opening ranges. As shown in Table~\ref{tab:related_benchmarks}, existing world model benchmarks can be grouped into two categories. One category targets general-purpose video generation, including VBench~\cite{huang2024vbench}, WorldModelBench~\cite{li2026worldmodelbench}, VideoPhy-2~\cite{bansal2026videophy}, and Physics-IQ~\cite{motamed2026generative}. These benchmarks offer broad evaluations of video quality, but may overlook robotics-specific physical details. The other category of benchmarks focuses on robotics, including RBench~\cite{deng2026rethinking} and the benchmarks introduced in ReWorld~\cite{peng2026reworld}, ABot-PhysWorld~\cite{chen2026abot}, and RoboGaze~\cite{nguyen2026robogaze}. These works emphasize physical plausibility in robotic scenarios. However, their evaluation protocols use general vision-language models (VLMs) without task-specific fine-tuning. Evaluating robotic manipulation videos requires robotics specific knowledge such as joint connections, ranges of motion, end-effector contact, and object support. General VLMs may lack certain knowledge. Meanwhile, physical violations, such as interpenetration and contact failures, can usually occur briefly and within small image regions. General VLMs without targeted training may miss these details, limiting evaluation accuracy. These limitations motivate a robotics-specific benchmark with a reward model specifically trained to accurately assess fine-grained physical plausibility of robotic manipulation videos.

\begin{table}[t]
    \centering
    \footnotesize
    \setlength{\tabcolsep}{0.8pt}
    \newcommand{\benchmarkhead}[1]{\raisebox{-.5\height}{\shortstack{#1}}}
    \caption{Comparison of WorldPhy with existing world model benchmarks.}
    \label{tab:related_benchmarks}
    \begin{tabular*}{\columnwidth}{@{\extracolsep{\fill}}lccc@{}}
        \toprule
        & \benchmarkhead{Robotics\\focus} & \benchmarkhead{Physical\\plausibility} & \benchmarkhead{Specifically trained\\reward model} \\
        \midrule
        VBench~\cite{huang2024vbench} & $\times$ & $\times$ & $\checkmark$ \\
        WorldModelBench~\cite{li2026worldmodelbench} & $\times$ & $\checkmark$ & $\checkmark$ \\
        VideoPhy-2~\cite{bansal2026videophy} & $\times$ & $\checkmark$ & $\checkmark$ \\
        Physics-IQ~\cite{motamed2026generative} & $\times$ & $\checkmark$ & $\times$ \\
        RBench~\cite{deng2026rethinking} & $\checkmark$ & $\checkmark$ & $\times$ \\
        ReWorld~\cite{peng2026reworld} & $\checkmark$ & $\checkmark$ & $\times$ \\
        ABot-PhysWorld~\cite{chen2026abot} & $\checkmark$ & $\checkmark$ & $\times$ \\
        RoboGaze~\cite{nguyen2026robogaze} & $\checkmark$ & $\checkmark$ & $\times$ \\
        \midrule
        \textbf{WorldPhy} & $\checkmark$ & $\checkmark$ & $\checkmark$ \\
        \bottomrule
    \end{tabular*}
\end{table}

We introduce WorldPhy, a benchmark tailored to robotics with a specifically trained reward model that provides accurate and fine-grained evaluation. We collect a large number of real-world and simulated videos of robotics scenarios, extract the corresponding images and instructions, and select 300 image--instruction pairs covering diverse viewpoints and tasks to evaluate world models. We also generate a large collection of videos through multiple world models and develop a detailed pipeline to annotate scores and corresponding reasons along multiple evaluation dimensions. After filtering, we obtain over 12K high-quality human-annotated examples and use them to train a dedicated reward model WorldPhyReward. Given the task inputs and a video generated by a world model, WorldPhyReward produces fine-grained scores and reasons across six physical evaluation dimensions and an instruction-following dimension.

Using this trained reward model, we evaluate 13 world models. The evaluation results show that Kling-3.0, MiniMax-H3, and Seedance2 lead in overall scores, but all models receive their two lowest scores for contact and interaction causality and for support and motion dynamics. Instruction following is generally a relative strength, yet no model achieves a perfect score on this dimension, and stronger instruction following does not necessarily imply better overall performance. Furthermore, 12 of the 13 models score lower in gripper scenarios than in hand scenarios, and most models receive their lowest scores under the wrist view among the three evaluated viewpoints. These findings indicate that, although some existing world models have achieved strong overall performance, they still cannot consistently ensure fine-grained physical plausibility across different robotics scenarios. Our contributions can be summarized as follows:

\begin{itemize}
    \item We construct a world model benchmark for robotics scenarios that supports fine-grained evaluation of physical plausibility.
    \item We develop a detailed pipeline to collect over 12K high-quality human-annotated video-scoring examples with fine-grained labels and train WorldPhyReward using this data. This model provides scores and reasons for each evaluation dimension and outperforms general-purpose VLMs such as GPT-6 Astra and video reward models such as WorldModelBench on our test set.
    \item We systematically evaluate 13 current open and closed-source world models. The results reveal shared weaknesses in contact causality and support dynamics, with additional limitations in gripper manipulation and wrist-view scenarios over these world models.
\end{itemize}
\section{Related Work}
\label{sec:related_work}

\noindent\textbf{Video World Models.}
Video world models learn spatiotemporal dynamics from videos and generate future frames conditioned on text, images, or other inputs to simulate object motion and interactions. Examples include Wan~\cite{wan2025wan}, Seedance 2.0~\cite{seedance2026seedance}, Cosmos 3~\cite{agarwal2026cosmos}, and MiniMax H3. These models generate videos with rich scene details and continuous dynamics under the given conditions, providing visual references for predicting future environmental states. In this paper, we focus on evaluation of video models that take in text and image conditions.

As video world models advance, an increasing number of studies apply them to robotics. Some use world models to generate training data, such as DreamGen~\cite{jang2025dreamgen} and GR00T N1~\cite{bjorck2025gr00t}, while others predict future frames to guide policy decisions, such as $\pi_{0.7}$~\cite{DBLP:journals/corr/abs-2604-15483} and UniPi~\cite{du2023learning}. These approaches improve the performance of embodied models. However, robotics tasks often place more demanding requirements on world models.

\noindent\textbf{Benchmarks of Video World Models.}
Most existing benchmarks evaluate world models broadly across general scenarios. VBench~\cite{huang2024vbench}, VMBench~\cite{ling2025vmbench}, and T2V-CompBench~\cite{sun2025t2v} assess overall video quality, motion quality, and compositionality, respectively. VBench-2.0~\cite{zheng2025vbench}, VideoPhy~\cite{bansal2025videophy}, VideoPhy-2~\cite{bansal2026videophy}, and PhyGenBench~\cite{meng2024towards} further evaluate physical laws and commonsense. Physics-IQ~\cite{motamed2026generative} tests video prediction through real-world physical experiments, while WorldScore~\cite{duan2025worldscore} assesses controllability, quality, and dynamics in multi-scene world generation. Other general-purpose benchmarks also include robotics tasks, such as WorldModelBench~\cite{li2026worldmodelbench}, PAI-Bench~\cite{zhou2026pai}, and WBench~\cite{ying2026wbench}.  These benchmarks provide broader coverage but can overlook the unique requirements of robotic world models.

Some benchmarks specifically target robotics. RBench~\cite{deng2026rethinking}, ReWorldBench from ReWorld~\cite{peng2026reworld}, EZSbench from ABot-PhysWorld~\cite{chen2026abot}, and RoboGaze~\cite{nguyen2026robogaze} evaluate generated videos by calling the APIs of general-purpose vision-language models (VLMs) without task-specific fine-tuning. Other studies, such as WoW-World-Eval~\cite{fan2026wow}, specifically train a VLM to assess physical commonsense, while also using many metrics based on ground-truth comparisons and pixel-level differences. However, the same instruction and initial frame can lead to multiple physically plausible motion sequences. Pixel or trajectory deviations from a single reference video therefore do not necessarily indicate physical violations and may cause valid alternative generations to be undervalued. Moreover, high visual similarity does not guarantee physically plausible local contact and object interactions, limiting the accuracy of these metrics for evaluation.

\noindent\textbf{Reward Models for Video Generation.}
A line of work trains VLMs to score generated videos. Examples include VideoReward~\cite{liu2026improving}, VisionReward~\cite{xu2026visionreward}, LiFT-Critic from LiFT~\cite{wang2024lift}, VideoScore2~\cite{he2025videoscore2}, and UnifiedReward-2.0~\cite{wang2025unified}. VideoReward and VisionReward learn to assess quality and human preferences across multiple dimensions, while LiFT-Critic and VideoScore2 combine score and rationale supervision. UnifiedReward-2.0 supports pairwise comparisons and per-dimension scoring across text alignment, physics, and style. These methods use human-annotated video scores to train scoring models aligned with human preferences. In this paper, we adopt a similar approach to train a reward model tailored to our benchmark.
\section{Method}
\label{sec:method}

\subsection{Data}

\subsubsection{Task Collection and Coverage}

To construct high-quality evaluation tasks with broad coverage of robotics scenarios, we collect 2,000 robot manipulation videos from 24 public robotics data repositories. These sources include VITRA-TeleData~\cite{li2026scalable}, the NVIDIA PhysicalAI series, such as GR00T-Teleop-GR1~\cite{gao2026dreamdojo} and GR00T-Teleop-G1, and datasets released by Pollen Robotics and Mimic Robotics. The videos primarily depict real-world robot manipulation and also include simulated scenarios.

We then extract the first frame of each video and provide it, along with the original video to GPT-5.6 Terra. The model generates an English instruction describing the robot's movement occured in the video. We manually verify and clean the generated descriptions and labels, ultimately obtaining 300 image--instruction pairs as a fixed set of evaluation tasks of our benchmark.

The task set covers three viewpoints (Third-person, Wrist, and Head), two end-effector types (Gripper and Hand), and two arm configurations (Single arm and Double arms). It includes ten task categories, such as Object Placement, Grasping, Stacking, and Cloth Folding, and five scene categories, such as Everyday tabletop, Kitchen, and Laboratory bench.

\subsubsection{Evaluation Dimensions}
\label{sec:evaluation_dimensions}

To accurately identify errors in videos generated by world models at sufficient granularity, we collect and inspect a large set of world-model-generated videos of robotics scenarios. We summarize the observed error types and organize them into seven evaluation dimensions, with representative violations shown in Figure~\ref{fig:worldphy_violations}. The first six assess fine-grained physical plausibility, while the seventh evaluates whether the generated video follows the instruction. These dimensions are defined as follows.

\textbf{Entity and Identity Consistency (EIC).} Detects unexplained changes in the presence, number, or identity of robotic arms and objects.

\textbf{Geometric and Material Plausibility (GMP).} Evaluates whether shapes, sizes, and deformations are consistent with material and structural properties.

\begin{figure}[!t]
    \centering
    \includegraphics[width=\columnwidth]{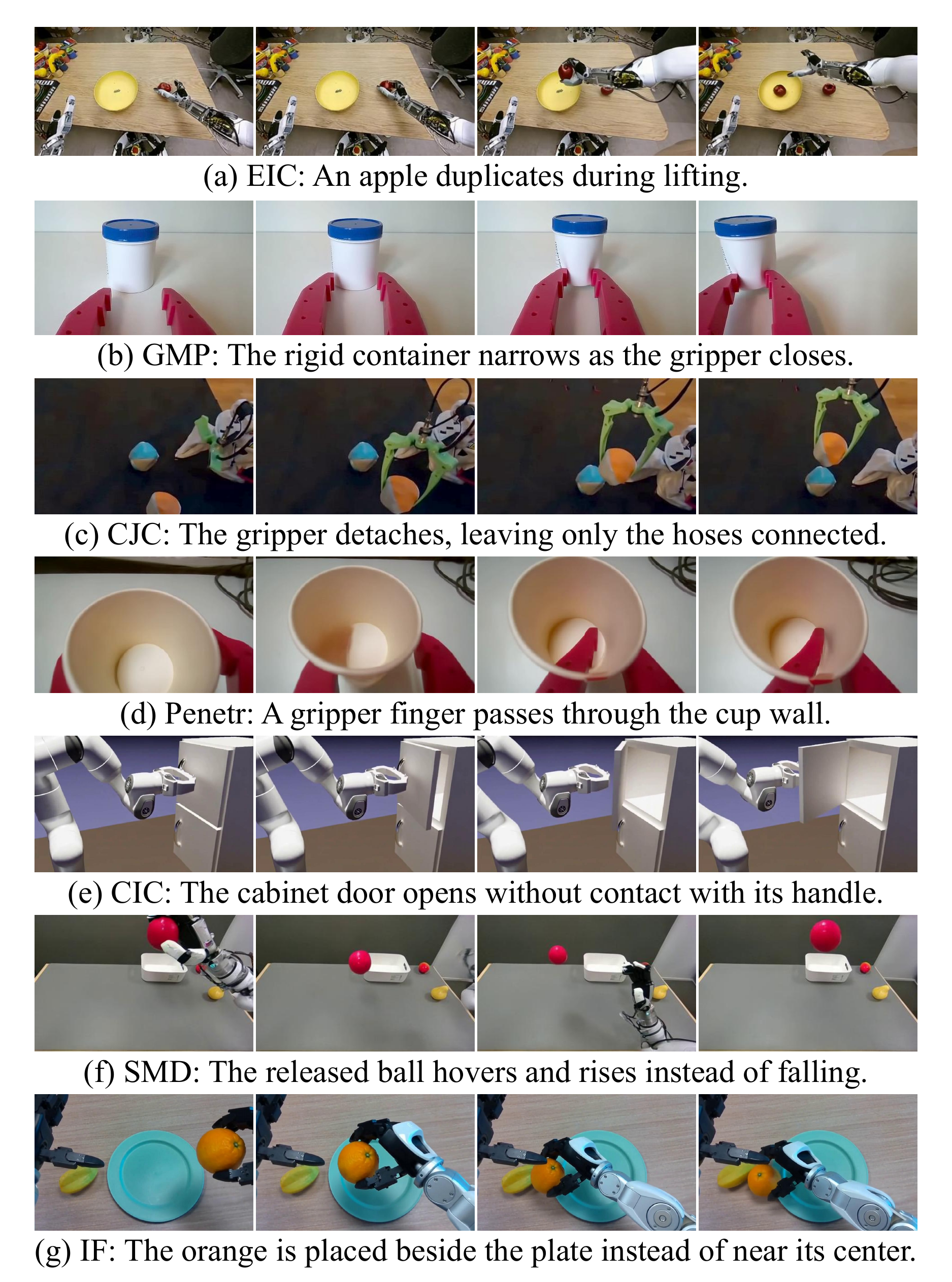}
    \ifnum\value{figure}=0\relax\setcounter{figure}{1}\fi
    \caption{Examples of violations across WorldPhy's evaluation dimensions.}
    \label{fig:worldphy_violations}
\end{figure}

\textbf{Connection and Joint Constraints (CJC).} Evaluates whether components maintain plausible connections and move within the constraints of their joints, hinges, or rails.

\textbf{Impenetrability (Penetr.).} Checks for physically implausible interpenetration between robotic arms, objects, and environmental surfaces.

\textbf{Contact and Interaction Causality (CIC).} Evaluates whether robot actions cause controlled object responses through valid contact or interaction pathways during grasping, pushing, pulling, and transport. Violations include an object being lifted while the gripper grasps empty space, moving with the robot before contact, or continuing to follow the gripper across a gap after release.

\textbf{Support and Motion Dynamics (SMD).} Evaluates whether object stability, motion, and collision responses are physically plausible given the existing support, forces, and motion state. Violations include unsupported hovering, falling too quickly or too slowly, and unexplained acceleration, stopping, changes in direction, or trajectory jumps.

\textbf{Instruction Following (IF).} Evaluates whether the manipulated objects, performed actions, and resulting states align with the input language instruction.

Each dimension receives a binary score: 1 if the corresponding criterion is satisfied, and 0 for violation. A video may violate multiple criteria simultaneously.

\subsubsection{Human Annotations}
\label{sec:human_annotations}

Training a high-quality reward model requires a large amount of high-quality training data. To obtain data of sufficient quantity and quality, we invite four annotators with experience in robotics to annotate the data.

First, we use the instructions and first-frame images from the 300 tasks as generation conditions. We provide these conditions to world models, including Cosmos 3 Nano, MiniMax H3, and Seedance 2.0, to generate approximately 2,000 robot manipulation videos of varying quality. Human annotators then assess each video against the six physical dimensions and the instruction-following dimension defined in Section~\ref{sec:evaluation_dimensions}, assigning a binary score to each dimension.

Physical violations in generated videos can be brief and occupy only a small region of the frame. Human scores alone may not clearly explain why a video violates a criterion. However, asking annotators to explain every violated dimension would be time-consuming and costly. After human scoring, we provide the videos and human ratings to GPT-6 Astra to generate reasons for dimensions labeled as violations. When GPT-6 Astra cannot explain a violation, annotators reassess the video. For items still judged to violate a criterion, they provide a brief description of the issue and an approximate timestamp. GPT-6 Astra then refines the explanation using these hints. Human annotators subsequently verify the explanations generated by GPT-6 Astra and discard incorrect items, resulting over 12K high-quality human ratings, each paired with an explanation.

From these over 12K human ratings, we hold out 1,400 ratings as a test set to evaluate our reward model's scoring quality. The test data cover outputs from Cosmos 3, MiniMax H3, and Seedance 2.0. Each evaluation dimension includes both compliant and violating samples, with coverage of different viewpoints and robot arm configurations. The remaining data are used to train our reward model.

\subsection{WorldPhyReward}

Using the high-quality human-annotated data in Section~\ref{sec:human_annotations}, we train our reward model, WorldPhyReward, to automatically score videos generated by world models. Given the conditioning image, task instruction, and candidate video, the model evaluates the six physical dimensions and the instruction-following dimension . For each dimension, WorldPhyReward first generates a reason and then outputs the corresponding binary score, learning to connect visible evidence in the video with its judgment.

\begin{figure*}[!t]
    \centering
    \includegraphics[width=\textwidth]{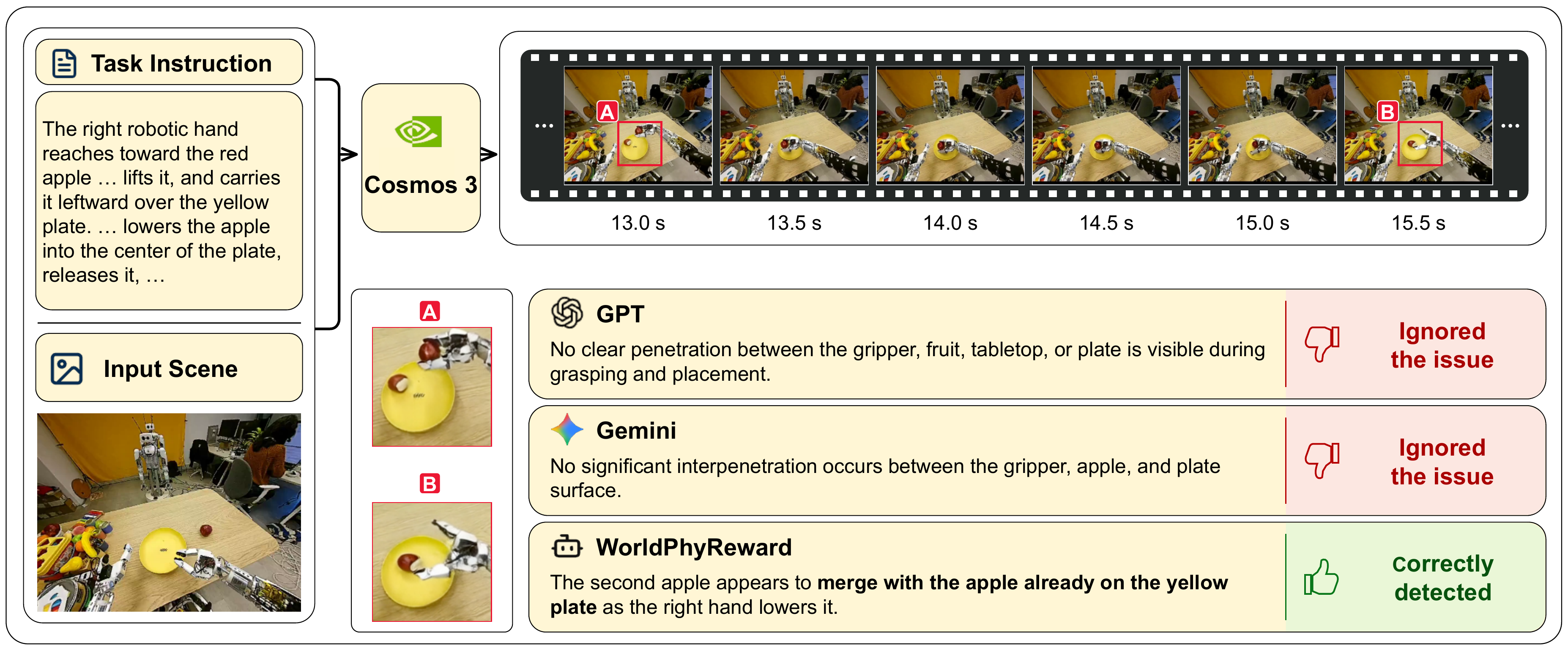}
    \caption{A comparison of WorldPhyReward with general VLMs. For a video generated from Cosmos 3, WorldPhyReward identifies and localizes a subtle physical violation missed by GPT and Gemini.}
    \label{fig:worldphyreward_demo}
\end{figure*}

During training, we observe that the model tends to overlook subtle violations. To strengthen learning from these errors, we adopt a class-weighted training objective that assigns a higher loss weight to output segments associated with annotated violations than to those associated with satisfied criteria. This weighting applies to both the reason and the score.

Let $x_i$ denote the input for video $i$, $y_i$ its target response, and $A_i$ the set of supervised answer-token positions. For a batch of $B$ videos, the objective is

\begin{equation}
\mathcal{L}(\theta)=-\frac{1}{B}\sum_{i=1}^{B}\frac{1}{|A_i|}\sum_{t\in A_i}w_{i,t}\log p_{\theta}(y_{i,t}\mid x_i,y_{i,<t}).
\end{equation}

Let $V_i\subseteq A_i$ contain the tokens in output segments for dimensions annotated as violated. The token weights are

\begin{equation}
w_{i,t}=\begin{cases}\lambda_{\mathrm{v}}, & t\in V_i,\\\lambda_{\mathrm{b}}, & \text{otherwise},\end{cases}\qquad \lambda_{\mathrm{v}}>\lambda_{\mathrm{b}}>0.
\end{equation}

The fixed weights $\lambda_{\mathrm{v}}$ and $\lambda_{\mathrm{b}}$ denote the violation and satisfied weights. Further training details are provided in Appendix~\ref{app:worldphyreward}.

This training yields a dedicated reward model for robotics scenarios that provides fine-grained scores and corresponding reasons across multiple evaluation dimensions. Figure~\ref{fig:worldphyreward_demo} illustrates its fine-grained identification of physical violations.
\begin{table*}[t]
\centering
\footnotesize
\setlength{\tabcolsep}{4pt}
\renewcommand{\arraystretch}{1.12}
\caption{Overall and per-dimension evaluation results of 13 world models on WorldPhy. Scores range from 0 to 1 for each dimension and from 0 to 7 overall, with higher values indicating better performance. Models are grouped as open or closed-source and sorted in descending order of overall score within each group. Bold indicates the highest value in each score column within a group.}
\label{tab:world_model_scores}
\begin{tabular*}{\textwidth}{@{\extracolsep{\fill}}lcccccccc@{}}
\toprule
\shortstack{\textbf{Model}} & \shortstack{\textbf{EIC}} & \shortstack{\textbf{GMP}} & \shortstack{\textbf{CJC}} & \shortstack{\textbf{Penetr.}} & \shortstack{\textbf{CIC}} & \shortstack{\textbf{SMD}} & \shortstack{\textbf{IF}} & \shortstack{\textbf{Mean total}\\\textbf{score}} \\
\midrule
\multicolumn{9}{l}{\textbf{Closed-source models}} \\
Kling 3.0 & \textbf{0.9356} & \textbf{0.9844} & \textbf{0.9967} & \textbf{0.9922} & 0.8511 & 0.8467 & \textbf{0.9900} & \textbf{6.5967} \\
Seedance 2.0 & 0.9158 & 0.9659 & 0.9932 & 0.9761 & \textbf{0.8601} & \textbf{0.8567} & 0.9841 & 6.5518 \\
\midrule
\multicolumn{9}{l}{\textbf{Open-source models}} \\
MiniMax H3 & \textbf{0.9733} & \textbf{0.9733} & \textbf{0.9900} & \textbf{0.9767} & \textbf{0.8300} & \textbf{0.8333} & \textbf{0.9856} & \textbf{6.5622} \\
Wan2.2-I2V-A14B & 0.9078 & 0.9333 & 0.9522 & 0.9378 & 0.7867 & 0.7867 & 0.9589 & 6.2633 \\
LongCat-Video & 0.8633 & 0.8967 & 0.9489 & 0.9122 & 0.7000 & 0.7000 & 0.9044 & 5.9256 \\
Kairos3.1-4B & 0.8700 & 0.8944 & 0.9500 & 0.9267 & 0.6867 & 0.6867 & 0.8144 & 5.8289 \\
HunyuanVideo-1.5 & 0.7311 & 0.8489 & 0.9511 & 0.8844 & 0.5956 & 0.5900 & 0.8667 & 5.4678 \\
Cosmos 3 & 0.6411 & 0.6800 & 0.8478 & 0.7233 & 0.5789 & 0.5722 & 0.8867 & 4.9300 \\
Wan2.2-TI2V-5B & 0.6811 & 0.7489 & 0.8178 & 0.7722 & 0.5556 & 0.5578 & 0.7622 & 4.8956 \\
WoW-1-Wan-14B & 0.7278 & 0.6844 & 0.7700 & 0.7156 & 0.5489 & 0.5489 & 0.7667 & 4.7622 \\
LTX-2.5 Distilled & 0.4289 & 0.5600 & 0.8233 & 0.6044 & 0.3900 & 0.3911 & 0.8856 & 4.0833 \\
LTX-2.5 Dev & 0.4200 & 0.4356 & 0.6767 & 0.5200 & 0.3378 & 0.3411 & 0.8311 & 3.5622 \\
CogVideoX1.5-5B & 0.1567 & 0.1089 & 0.2278 & 0.1467 & 0.0889 & 0.0833 & 0.2867 & 1.0989 \\
\bottomrule
\end{tabular*}
\end{table*}

\section{Experiments}
\label{sec:experiments}

This section evaluates both the generation performance of world models and the scoring quality of our reward model. First, we evaluate 13 open and closed-source world models using our benchmark. We use our trained reward model to score the generated videos. We compare overall scores and scores for the six physical plausibility dimensions and the instruction-following dimension. The comparison also covers performance across different end effectors, robot arm configurations, and camera viewpoints. We then compare our trained reward model with open and closed-source general VLMs and existing video reward models to assess its agreement with human judgments.

\begin{table*}[t]
\centering
\footnotesize
\setlength{\tabcolsep}{4pt}
\renewcommand{\arraystretch}{1.12}
\caption{Mean overall scores of 13 world models across different categories of robotics scenarios. Scores range from 0 to 7, with higher values indicating better performance.}
\label{tab:world_model_categories}
\begin{tabular*}{\textwidth}{@{\extracolsep{\fill}}lccccccc@{}}
\toprule
\textbf{Category} & \multicolumn{2}{c}{\textbf{End effector}} & \multicolumn{2}{c}{\textbf{Arm configuration}} & \multicolumn{3}{c}{\textbf{Camera viewpoint}} \\
\cmidrule(lr){2-3}\cmidrule(lr){4-5}\cmidrule(lr){6-8}
\shortstack{\textbf{Model}} & \shortstack{\textbf{Gripper}} & \shortstack{\textbf{Hand}} & \shortstack{\textbf{Single}\\\textbf{arm}} & \shortstack{\textbf{Double}\\\textbf{arms}} & \shortstack{\textbf{Third-person}} & \shortstack{\textbf{Wrist}} & \shortstack{\textbf{Head}} \\
\midrule
\multicolumn{8}{l}{\textbf{Closed-source models}} \\
Kling 3.0 & \textbf{6.3667} & 6.8267 & \textbf{6.4911} & 6.7022 & \textbf{6.7533} & \textbf{6.2600} & 6.7767 \\
Seedance 2.0 & 6.2315 & \textbf{6.8613} & 6.3356 & \textbf{6.7635} & 6.4867 & 6.2199 & \textbf{6.9327} \\
\midrule
\multicolumn{8}{l}{\textbf{Open-source models}} \\
MiniMax H3 & \textbf{6.1756} & \textbf{6.9489} & \textbf{6.5756} & \textbf{6.5489} & \textbf{6.7533} & \textbf{5.9900} & \textbf{6.9433} \\
Wan2.2-I2V-A14B & 5.7867 & 6.7400 & 6.0378 & 6.4889 & 6.2000 & 5.8400 & 6.7500 \\
LongCat-Video & 5.5289 & 6.3222 & 6.0511 & 5.8000 & 6.0100 & 5.5900 & 6.1767 \\
Kairos3.1-4B & 5.2422 & 6.4156 & 5.5867 & 6.0711 & 6.0367 & 4.9967 & 6.4533 \\
HunyuanVideo-1.5 & 5.2089 & 5.7267 & 5.4178 & 5.5178 & 5.2633 & 5.5367 & 5.6033 \\
Cosmos 3 & 4.3756 & 5.4844 & 4.4844 & 5.3756 & 4.7567 & 4.4700 & 5.5633 \\
Wan2.2-TI2V-5B & 4.7511 & 5.0400 & 4.8711 & 4.9200 & 4.6333 & 5.2100 & 4.8433 \\
WoW-1-Wan-14B & 4.3756 & 5.1489 & 4.3422 & 5.1822 & 4.3333 & 4.3933 & 5.5600 \\
LTX-2.5 Distilled & 3.7867 & 4.3800 & 4.0444 & 4.1222 & 3.7400 & 4.3100 & 4.2000 \\
LTX-2.5 Dev & 3.3622 & 3.7622 & 3.3756 & 3.7489 & 3.1700 & 3.6567 & 3.8600 \\
CogVideoX1.5-5B & 1.4556 & 0.7422 & 1.4267 & 0.7711 & 0.6367 & 1.8000 & 0.8600 \\
\bottomrule
\end{tabular*}
\end{table*}

\subsection{Evaluation results}

\begin{table*}[t]
\centering
\scriptsize
\setlength{\tabcolsep}{1.5pt}
\renewcommand{\arraystretch}{1.12}
\caption{Per-dimension scoring performance of WorldPhyReward and general-purpose VLMs on 1,400 human rubric ratings. We evaluate F1 score and accuracy. Higher values are better. For invalid responses generated by different models, F1 and accuracy use different conventions. See Appendix~\ref{app:table4_evaluation_protocol} for definitions and an extended confusion matrix.}
\label{tab:vlm_quality}
\begin{tabular*}{\textwidth}{@{\extracolsep{\fill}}l*{17}{c}@{}}
\toprule
\textbf{Model} & \multicolumn{2}{c}{\textbf{EIC}} & \multicolumn{2}{c}{\textbf{GMP}} & \multicolumn{2}{c}{\textbf{CJC}} & \multicolumn{2}{c}{\textbf{Penetr.}} & \multicolumn{2}{c}{\textbf{CIC}} & \multicolumn{2}{c}{\textbf{SMD}} & \multicolumn{2}{c}{\textbf{IF}} & \multicolumn{3}{c}{\textbf{Overall}} \\
\cmidrule(lr){2-3}\cmidrule(lr){4-5}\cmidrule(lr){6-7}\cmidrule(lr){8-9}\cmidrule(lr){10-11}\cmidrule(lr){12-13}\cmidrule(lr){14-15}\cmidrule(lr){16-18}
 & \shortstack{F1\\(\%)} & \shortstack{Acc.\\(\%)} & \shortstack{F1\\(\%)} & \shortstack{Acc.\\(\%)} & \shortstack{F1\\(\%)} & \shortstack{Acc.\\(\%)} & \shortstack{F1\\(\%)} & \shortstack{Acc.\\(\%)} & \shortstack{F1\\(\%)} & \shortstack{Acc.\\(\%)} & \shortstack{F1\\(\%)} & \shortstack{Acc.\\(\%)} & \shortstack{F1\\(\%)} & \shortstack{Acc.\\(\%)} & \shortstack{Micro\\F1 (\%)} & \shortstack{Macro\\F1 (\%)} & \shortstack{Acc.\\(\%)} \\
\midrule
Qwen3.8-27B & 7.79 & 7.50 & 9.38 & 4.83 & 20.51 & 8.50 & 15.63 & 6.17 & 20.71 & 6.67 & 22.31 & 7.17 & 13.77 & 7.00 & 16.05 & 15.73 & 6.83 \\
Claude Opus 5.5 & 64.91 & 48.00 & 62.85 & 43.83 & 72.52 & 53.67 & 71.08 & 51.50 & 64.57 & 47.00 & 64.08 & 44.17 & 64.59 & 48.50 & 66.67 & 66.37 & 48.10 \\
Seed2.1 Pro & 69.49 & 57.83 & 66.57 & 51.17 & 81.78 & 68.67 & 58.78 & 55.83 & 67.17 & 58.00 & 66.76 & 55.50 & 69.49 & 62.67 & 69.36 & 68.58 & 58.52 \\
Qwen3.5-27B & 67.81 & 52.17 & 66.97 & 50.33 & 82.06 & 69.50 & 78.45 & 64.00 & 65.26 & 50.17 & 66.05 & 50.33 & 74.60 & 63.33 & 71.94 & 71.60 & 57.12 \\
Qwen3.6-27B & 68.34 & 53.67 & 66.67 & 50.00 & 82.35 & 70.00 & 78.05 & 64.00 & 66.96 & 50.67 & 66.89 & 50.50 & 76.29 & 65.50 & 72.49 & 72.22 & 57.76 \\
GPT-5.6 Sol & 72.12 & 61.83 & 68.66 & 55.50 & 82.30 & 70.33 & 78.58 & 65.17 & 68.16 & 53.83 & 67.79 & 53.83 & 76.42 & 69.00 & 73.72 & 73.43 & 61.36 \\
GPT-6 Astra & 77.11 & 71.50 & 73.31 & 65.17 & 84.04 & 73.67 & 78.96 & 66.17 & 70.43 & 59.83 & 71.17 & 60.83 & 75.49 & \textbf{70.67} & 76.12 & 75.79 & 66.83 \\
Gemini 3.8 Flash & 75.41 & 70.50 & 77.38 & 72.50 & 84.58 & 76.00 & 80.33 & 72.33 & \textbf{73.59} & \textbf{69.67} & \textbf{73.36} & \textbf{69.00} & 73.37 & 67.67 & 77.27 & 76.86 & 71.10 \\
\midrule
\textbf{WorldPhyReward (ours)} & \textbf{78.83} & \textbf{75.83} & \textbf{79.88} & \textbf{76.67} & \textbf{86.19} & \textbf{79.33} & \textbf{84.22} & \textbf{78.83} & 70.63 & 69.50 & 66.88 & 65.83 & \textbf{79.27} & 69.83 & \textbf{78.75} & \textbf{77.99} & \textbf{73.69} \\
\bottomrule
\end{tabular*}
\end{table*}
\begin{table}[t]
\centering
\scriptsize
\setlength{\tabcolsep}{3.0pt}
\renewcommand{\arraystretch}{1.12}
\caption{Correlation with human physical-plausibility scores summed over six dimensions. Pearson measures linear correlation and Spearman measures rank correlation.}
\label{tab:reward_quality}
\begin{tabular*}{\columnwidth}{@{\extracolsep{\fill}}lcc@{}}
\toprule
\textbf{Model} & \textbf{Pearson} $r\uparrow$ & \textbf{Spearman} $\rho\uparrow$ \\
\midrule
VideoCon-Physics & 0.0759 & 0.0546 \\
VideoReward & 0.0849 & 0.0495 \\
VideoPhy-2-AutoEval & 0.0907 & 0.0952 \\
VideoScore2 & 0.3001 & 0.2976 \\
WorldModelBench & 0.3042 & 0.2811 \\
UnifiedReward-2.0 & 0.3122 & 0.3002 \\
RoboJudge-9B & 0.4266 & 0.4314 \\
\midrule
\textbf{WorldPhyReward (ours)} & \textbf{0.6169} & \textbf{0.5919} \\
\bottomrule
\end{tabular*}
\end{table}

We use WorldPhy's 300 tasks as shared generation conditions to evaluate 11 open-source models, including MiniMax H3, Wan2.2~\cite{wan2025wan}, LongCat-Video, Kairos3.1-4B, HunyuanVideo-1.5~\cite{wu2025hunyuanvideo}, Cosmos 3, WoW-1-Wan-14B~\cite{chi2025wow}, LTX-2.5~\cite{hacohen2026ltx}, and CogVideoX1.5-5B~\cite{yang2025cogvideox}, and two closed-source models, Kling 3.0 and Seedance 2.0. Our trained reward model takes the instruction, first-frame image, and generated video as input and assigns a score to each of the seven evaluation dimensions. We average scores over three runs and report the mean for each category. The total score is the sum of the seven metric scores. Table~\ref{tab:world_model_scores} summarizes the overall and per-dimension scores, while Table~\ref{tab:world_model_categories} further reports performance across end effectors, robot arm configurations, and camera viewpoints.

As shown in Table~\ref{tab:world_model_scores}, Kling 3.0, MiniMax H3, and Seedance 2.0 achieve the three highest overall scores of 6.5967, 6.5622, and 6.5518, respectively. Although these leading models achieve high overall scores, the individual dimensions and scenario categories still reveal limitations in their physical plausibility.

In Table~\ref{tab:world_model_scores}, CIC and SMD are the two lowest-scoring dimensions for all 13 models. Even for the top three models overall, scores on these dimensions range only from 0.8300 to 0.8601. This indicates that current models still struggle to reproduce plausible robot--object interactions and object motion under the given support and force conditions. IF ranks among the top three dimensions for 12 of the 13 models, but none achieves a perfect IF score. Moreover, a higher IF score does not always correspond to a higher overall score. For example, LTX-2.5 Distilled achieves a higher IF score than HunyuanVideo-1.5 (0.8856 vs. 0.8667), but a lower overall score (4.0833 vs. 5.4678). Thus, depicting the actions and outcomes specified by an instruction is insufficient to ensure fine-grained physical plausibility throughout the manipulation process.

\begingroup\widowpenalty=10000 Overall averages alone cannot fully characterize model performance, category-level analyses are also needed to identify limitations. As shown in Table~\ref{tab:world_model_categories}, 12 of the 13 models score lower in Gripper scenarios than in Hand scenarios, and the top six models overall obtain their lowest scores among the three viewpoints in Wrist views. For example, MiniMax H3 achieves an overall score of 6.5622, but scores only 5.9900 in Wrist views, compared with 6.9433 in Head views. These results indicate that, despite strong overall performance, existing world models still struggle to consistently maintain fine-grained physical plausibility across different robotics scenarios.\par\endgroup

\subsection{Reward Model Quality}
\label{sec:reward_model_quality}

We evaluate WorldPhyReward on a test set of 1,400 human rubric ratings that were not used for training. The test data covers outputs from Cosmos 3, MiniMax H3, and Seedance 2.0, and includes both compliant and violating samples in each evaluation dimension. It contains balanced label distributions and coverage of different viewpoints and robot arm configurations. The comparative evaluations use three different seeds under the same configuration; we report the mean across the three runs.

We assess agreement with human judgments at both the per-dimension and overall score levels. Table~\ref{tab:vlm_quality} compares WorldPhyReward with open and closed-source general-purpose VLMs, including Qwen3.8-27B, Claude Opus 5.5, Seed2.1 Pro, Qwen3.5-27B, Qwen3.6-27B, GPT-5.6 Sol, GPT-6 Astra, and Gemini 3.8 Flash, reporting per-dimension F1 and Accuracy as well as aggregate metrics. Table~\ref{tab:reward_quality} compares WorldPhyReward with existing video reward models, including VideoCon-Physics~\cite{bansal2025videophy}, VideoReward~\cite{liu2026improving}, VideoPhy-2-AutoEval~\cite{bansal2026videophy}, VideoScore2~\cite{he2025videoscore2}, WorldModelBench~\cite{li2026worldmodelbench}, UnifiedReward-2.0~\cite{wang2025unified}, and RoboJudge-9B~\cite{ma2026robojudge}, using Pearson and Spearman correlations with the summed human scores across the six physical dimensions. WorldPhyReward sums its scores across the same six dimensions, while the baselines use their respective overall physical-plausibility scores. VideoReward uses MQ+VQ score, and WorldModelBench uses only its physical component score. Each model retains its original input and output processing logic.

As shown in Tables~\ref{tab:vlm_quality} and~\ref{tab:reward_quality}, WorldPhyReward achieves the highest Macro F1, Micro F1, and overall Accuracy of $77.99\%$, $78.75\%$, and $73.69\%$, respectively. The mean scores improve over the strongest general-purpose VLM baseline, Gemini 3.8 Flash, by 1.13, 1.48, and 2.59 percentage points. WorldPhyReward also achieves the highest mean F1 in five dimensions and the highest mean Accuracy in four dimensions. Although WorldPhyReward slightly trails some closed-source VLMs on individual CIC, SMD, and IF dimensions, it remains competitive in these dimensions. Its stronger overall performance and local deployment make these modest gaps a reasonable trade-off for large-scale evaluation without costly proprietary API calls. Its physical-score Pearson correlation reaches $0.6169$, compared with $0.4266$ for the strongest reward-model baseline, RoboJudge-9B. Its Spearman correlation reaches $0.5919$, compared with $0.4314$ for the same baseline. Overall, WorldPhyReward shows greater agreement with human judgments in aggregate classification, physical scoring, and ranking.

We also compare WorldPhyReward with existing reward-model baselines in assessing instruction following. Details and results are provided in Appendix~\ref{app:instruction_following}.

\subsection{Human Alignment}
\label{sec:human_alignment}

\begin{table*}[t]
\centering
\caption{Human preference alignment results. (a) Evaluation data generated by world models represented in WorldPhyReward's training set. (b) Evaluation data generated by world models absent from WorldPhyReward's training set.}
\label{tab:human_alignment}
\begin{minipage}[t]{0.48\textwidth}
\centering
{\small\normalfont (a) Seen generators\par}\smallskip
\scriptsize
\setlength{\tabcolsep}{3pt}
\renewcommand{\arraystretch}{1.12}
\begin{tabular*}{\linewidth}{@{\extracolsep{\fill}}lcc@{}}
\toprule
\textbf{Reward model} & \textbf{Correct / 50} & \textbf{Accuracy (\%)} \\
\midrule
VideoCon-Physics & 20/50 & 40.00 \\
UnifiedReward-2.0 & 20/50 & 40.00 \\
VideoPhy-2-AutoEval & 22/50 & 44.00 \\
VideoReward & 22/50 & 44.00 \\
VideoScore2 & 23/50 & 46.00 \\
RoboJudge-9B & 24/50 & 48.00 \\
WorldModelBench & 25/50 & 50.00 \\
\midrule
\textbf{WorldPhyReward (ours)} & \textbf{39/50} & \textbf{78.00} \\
\bottomrule
\end{tabular*}
\end{minipage}\hfill
\begin{minipage}[t]{0.48\textwidth}
\centering
{\small\normalfont (b) Unseen generators\par}\smallskip
\scriptsize
\setlength{\tabcolsep}{3pt}
\renewcommand{\arraystretch}{1.12}
\begin{tabular*}{\linewidth}{@{\extracolsep{\fill}}lcc@{}}
\toprule
\textbf{Reward model} & \textbf{Correct / 50} & \textbf{Accuracy (\%)} \\
\midrule
VideoCon-Physics & 26/50 & 52.00 \\
UnifiedReward-2.0 & 27/50 & 54.00 \\
VideoPhy-2-AutoEval & 17/50 & 34.00 \\
VideoReward & 17/50 & 34.00 \\
VideoScore2 & 22/50 & 44.00 \\
RoboJudge-9B & 28/50 & 56.00 \\
WorldModelBench & 26/50 & 52.00 \\
\midrule
\textbf{WorldPhyReward (ours)} & \textbf{38/50} & \textbf{76.00} \\
\bottomrule
\end{tabular*}
\end{minipage}
\end{table*}

To further evaluate whether WorldPhyReward aligns with human preferences, we collect 50 pairs of generated videos corresponding to 50 distinct combinations of task instructions and initial frames. The two videos in each pair share the same generation conditions. All 100 videos are unique, and each appears in only one pair. The videos come from Cosmos 3, MiniMax H3, and Seedance 2.0, with 30 pairs comparing outputs from different models and 20 pairs comparing different outputs from the same model. These conditions are selected from WorldPhy's 300 tasks and cover grasping and transport, pose adjustment, component fitting, and stacking.

Human annotators then indicate their preference for one video in each pair or select no preference. WorldPhyReward scores each video separately and prefers the one with the higher total score. We measure alignment with human preferences by the accuracy of these preference predictions against the human labels. For comparison, we apply the same procedure to the overall scores of VideoCon-Physics, VideoPhy-2-AutoEval, VideoReward, VideoScore2, UnifiedReward-2.0, WorldModelBench, and RoboJudge-9B.

As shown in Table~\ref{tab:human_alignment}(a), WorldPhyReward correctly predicts human preferences for 39 of the 50 pairs, achieving an accuracy of 78.00\% and outperforming all baselines. The strongest baseline, WorldModelBench, achieves 50.00\% accuracy, which WorldPhyReward exceeds by 28 percentage points. The remaining baselines achieve accuracies of 40.00\%--48.00\%. These results indicate that WorldPhyReward not only provides fine-grained scores but also produces overall scores that more accurately reflect human preferences for generated videos.

We further evaluate whether WorldPhyReward maintains strong alignment with human preferences on videos generated by world models absent from its training data. Following the same protocol, we collect 50 video pairs generated by world models other than Cosmos 3, MiniMax H3, and Seedance 2.0, including Kling 3.0 and HunyuanVideo. Human annotators indicate their preferences for these video pairs. We then score the videos using WorldPhyReward and the baseline models, infer each model's preferences from its total scores, and compute preference prediction accuracy.

As shown in Table~\ref{tab:human_alignment}(b), WorldPhyReward correctly predicts human preferences for 38 of the 50 pairs, achieving an accuracy of 76.00\% and outperforming all comparison models. The strongest baseline, RoboJudge-9B, achieves 56.00\% accuracy, which WorldPhyReward exceeds by 20 percentage points. Compared with its 78.00\% accuracy on seen generators, the accuracy decreases by only 2 percentage points, indicating that WorldPhyReward maintains strong alignment with human preferences on unseen generators.

Overall, WorldPhyReward accurately predicts human preferences for videos from both seen and unseen generators and outperforms the comparison models, demonstrating strong generalization to unseen generators.
\section{Conclusion}
\label{sec:conclusion}

This paper introduces WorldPhy to benchmark the fine-grained physical plausibility of video world models in robotics. Existing evaluations can overlook robotics-specific physical details and struggle to assess them accurately. WorldPhy provides 300 diverse image--instruction tasks and evaluates generated videos across six physical dimensions and an instruction-following dimension. We collect over 12K human-annotated dimension scores with accompanying explanations and fine-tune WorldPhyReward to automatically provide scores and reasons. Comparative evaluations show that WorldPhyReward better aligns with human ratings and video preferences than the corresponding baselines. Our evaluation of 13 world models reveals persistent weaknesses in contact causality and support dynamics, with further limitations in gripper manipulation and wrist-view scenarios. Through providing fine-grained evaluations over physical plausibility on robotics scenarios, WorldPhy can support the development of more physically plausible world models and their reliable use in robotics. Future work will explore using WorldPhyReward to guide world-model training and improve downstream robotic control.
\clearpage
\section*{LLM Usage Statement}

We used GPT-5.6 Terra to draft task instructions from video frames, label camera and robot configurations, and classify tasks by action. Human annotators scored videos from Cosmos 3 Nano, MiniMax H3, and Seedance 2.0. GPT-6 Astra drafted rationales for those scores and revised them using annotator feedback. We fine-tuned Qwen3.5-9B on these annotations to obtain WorldPhyReward. WorldPhyReward then scored videos from 13 world models, including videos generated via the Kling 3.0 and Seedance 2.0 APIs. For comparison with human ratings, we also scored the test videos using Claude Opus 5.5, Seed2.1 Pro, GPT-5.6 Sol, GPT-6 Astra, and Gemini 3.8 Flash. AI assistants (ChatGPT and Codex) helped us write and debug code, check experimental results, translate and edit the manuscript, and format tables. The authors take full responsibility for the manuscript and its reported results.
{\small
  \phantomsection
  \pdfbookmark[1]{References}{preprint.references}

}

\clearpage
\raggedbottom
\appendixfloatlayout
\maketitlesupplementary
\pdfbookmark[1]{Appendix Contents}{preprint.appendixcontents}
\addtocontents{toc}{\protect\setcounter{tocdepth}{2}}
\begin{strip}
  \appendixcontents
\end{strip}
\clearpage
\appendix
\begin{table}[t]
\centering
\footnotesize
\setlength{\tabcolsep}{3pt}
\renewcommand{\arraystretch}{1.1}
\caption{WorldPhyReward training configuration.}
\label{tab:training_configuration}
\begin{tabular}{@{}p{.35\columnwidth}p{.59\columnwidth}@{}}
\toprule
\textbf{Parameter} & \textbf{Setting} \\
\midrule
Frame sampling
  & Uniform over the full video, target 5 fps, up to 128 frames \\
Initial image and video frames
  & RGB, bicubic resize to $640\times480$, no crop \\
Trainable parameters
  & All 9.41B, including vision backbone and multimodal projector \\
Hardware and precision
  & 4 $\times$ A800 (80 GB each), BF16 mixed precision \\
Distributed training
  & ZeRO-3 without CPU or NVMe offloading \\
Batch size
  & 1 per GPU, 8 accumulation steps, 32 globally \\
Optimizer
  & AdamW ($\beta_1=0.9$, $\beta_2=0.999$, $\epsilon=10^{-8}$) \\
Weight decay
  & 0.01 \\
Gradient clipping
  & Maximum gradient norm of 1.0 \\
Learning rates
  & Language: $3\times10^{-6}$, vision: $3\times10^{-8}$ \\
training steps& 200 steps (4 epochs) \\
\bottomrule
\end{tabular}
\end{table}

\section{WorldPhyReward Training Details and Loss-Weight Ablation}
\label{app:worldphyreward}

\subsection{Training Details}
\label{app:training_details}

\paragraph{Model and supervision.}
We fine-tune all parameters of Qwen3.5-9B using LLaMA-Factory.
Human annotators assigned seven binary ratings to each of the training videos, one per evaluation dimension.
To obtain an English rationale for violation rating, we provide GPT-6 Astra with video frames sampled at 10 fps.
When a rationale fails to identify the violation, an annotator reviews the video and supplies a description and an approximate timestamp to guide the revision.

\paragraph{Inputs and outputs.}
WorldPhyReward takes the initial image, task instruction, candidate video, and definitions of the seven evaluation dimensions as input.
It generates rationales and scores directly as a JSON object without a separate scalar reward head.
Fields \texttt{R1} through \texttt{R7} correspond, in order, to EIC, GMP, CJC, Penetr., CIC, SMD, and IF.
For each field, the model generates an English rationale followed by a binary score.
A rationale beginning with \texttt{Issue:} describes the observed violation and is followed by a score of 0.
A rationale beginning with \texttt{Consistent:} describes how the video satisfies the criterion and is followed by a score of 1.
The prompt limits each rationale to 45 English words, including the prefix.
Training and inference use the same prompt and visual preprocessing.

\paragraph{Loss computation.}
We compute loss only on the target JSON tokens (field names, rationales, binary scores, and formatting) and the end-of-sequence token.
Each dimension's span ends at its score token, inclusive.
If the human label for that dimension is 0, all tokens in the span receive a weight of 2.
All other supervised tokens receive a weight of 1, including the closing JSON tokens and the end-of-sequence token after the final score.
For each response, we divide the sum of weighted token losses by the number of supervised tokens, then average the resulting losses across the batch.

\begin{table}[t]
\centering
\footnotesize
\setlength{\tabcolsep}{3pt}
\renewcommand{\arraystretch}{1.1}
\caption{Ablation result}
\label{tab:loss_weight_ablation}
\begin{tabular}{@{}p{.47\columnwidth}rr@{}}
\toprule
\textbf{Metric}
  & \shortstack{Uniform\\(1:1)}
  & \shortstack{WorldPhyReward\\(2:1)} \\
\midrule
Accuracy (\%)                                 & 66.79  & 73.69  \\
Micro F1 (\%)                                 & 76.62  & 78.75  \\
Macro F1 (\%)                                 & 76.20  & 77.99  \\
\bottomrule
\end{tabular}
\end{table}

\subsection{Loss-Weight Ablation}
\label{app:loss_ablation}

\paragraph{Training setup.}
We compare WorldPhyReward, which uses a 2:1 weight ratio for violated versus satisfied criteria, with a uniformly weighted baseline (1:1).
Both models start from the same Qwen3.5-9B checkpoint and use the same training data and settings in Table~\ref{tab:training_configuration}.
The only change to the training objective is the weight assigned to spans labeled as violations.

\paragraph{Inference setup.}
Both models use the same prompt, output parser, and decoding settings.
We run vLLM in BF16 with tensor parallelism across two A800 GPUs, a batch size of 16, a context limit of 24,576 tokens, and a maximum of 1,024 new tokens per response.
We set the temperature to 0 and top-$p$ to 1, with thinking mode and additional structured-output repair disabled.

\paragraph{Evaluation set and reporting.}
The evaluation set contains 200 videos with 1,400 human ratings.
Of these, 610 indicate a violation and 790 indicate that the criterion is satisfied.
Table~\ref{tab:loss_weight_ablation} provides a descriptive comparison on this set.

For F1, the positive class is a satisfied criterion (score 1).
We compute micro F1 by pooling judgments across all seven dimensions and macro F1 by averaging the seven dimension-wise F1 scores.

\clearpage
\begin{table}[t]
\centering
\footnotesize
\renewcommand{\thetable}{\arabic{table}(a)}
\renewcommand{\arraystretch}{1.15}
\caption{Extended confusion matrix for one evaluation dimension in one run.}
\label{tab:invalid_response_matrix}
\begin{tabular*}{\columnwidth}{@{\extracolsep{\fill}}lccc@{}}
\toprule
\textbf{Human label} & \textbf{Predicted 1} & \textbf{Predicted 0} & \textbf{Invalid} \\
\midrule
1 (satisfied) & $\mathrm{TP}$ & $\mathrm{FN}$ & $u_1$ \\
0 (violation) & $\mathrm{FP}$ & $\mathrm{TN}$ & $u_0$ \\
\bottomrule
\end{tabular*}
\end{table}

\subsection{Evaluation Protocol for general VLMs}
\label{app:table4_evaluation_protocol}

\paragraph{Invalid responses and metric definitions.}
If a response cannot be parsed, its predictions for all seven dimensions are marked as invalid.
Table~\ref{tab:invalid_response_matrix} shows how we count predictions for each dimension.
The first two prediction columns contain valid responses. The counts $u_1$ and $u_0$ denote invalid responses with human labels 1 and 0, respectively.
For $F_1$, label 1 is positive. Each of the $u_1$ invalid responses counts as a false negative, while the $u_0$ invalid responses are ignored in the $F_1$ calculation.
Accuracy counts every invalid prediction as incorrect and uses all $N=200$ responses in the run.
\[
\begin{array}{@{}c@{\;}c@{\;}c@{}}
  F_1 & = & \displaystyle\frac{2\mathrm{TP}}{2\mathrm{TP}+\mathrm{FP}+\mathrm{FN}+u_1} \\[0.75em]
  \mathrm{Accuracy} & = & \displaystyle\frac{\mathrm{TP}+\mathrm{TN}}{N}
\end{array}
\]

\newpage
\begin{table}[t]
\ContinuedFloat
\renewcommand{\thetable}{\arabic{table}(b)}
\centering
\footnotesize
\setlength{\tabcolsep}{3pt}
\renewcommand{\arraystretch}{1.12}
\caption{Successfully parsed responses in the main-paper VLM evaluation. Each run contains 200 responses; the overall valid-output rate is computed over all 600 responses.}
\label{tab:valid_output_rates}
\begin{tabular*}{\columnwidth}{@{\extracolsep{\fill}}lrrrr@{}}
\toprule
\textbf{Model} & \textbf{Run 1} & \textbf{Run 2} & \textbf{Run 3} & \shortstack{\textbf{Overall}\\\textbf{valid (\%)}} \\
\midrule
Qwen3.8-27B & 23/200 & 27/200 & 27/200 & 12.83 \\
Claude Opus 5.5 & 144/200 & 154/200 & 150/200 & 74.67 \\
Seed2.1 Pro & 187/200 & 199/200 & 200/200 & 97.67 \\
Qwen3.5-27B & 199/200 & 198/200 & 198/200 & 99.17 \\
Qwen3.6-27B & 200/200 & 200/200 & 200/200 & 100.00 \\
GPT-5.6 Sol & 196/200 & 198/200 & 198/200 & 98.67 \\
GPT-6 Astra & 200/200 & 200/200 & 200/200 & 100.00 \\
Gemini 3.8 Flash & 198/200 & 200/200 & 200/200 & 99.67 \\
\midrule
WorldPhyReward (ours) & 200/200 & 200/200 & 200/200 & 100.00 \\
\bottomrule
\end{tabular*}
\end{table}

\paragraph{Valid-output statistics.}
Table~\ref{tab:valid_output_rates} reports the number of successfully parsed responses in each run and the overall valid-output rate across the three runs.
\clearpage
\twocolumn[{
\begin{minipage}{\textwidth}
\captionsetup{hypcap=false}
\centering
\includegraphics[width=\dimexpr(\textwidth-\columnsep)/2\relax]{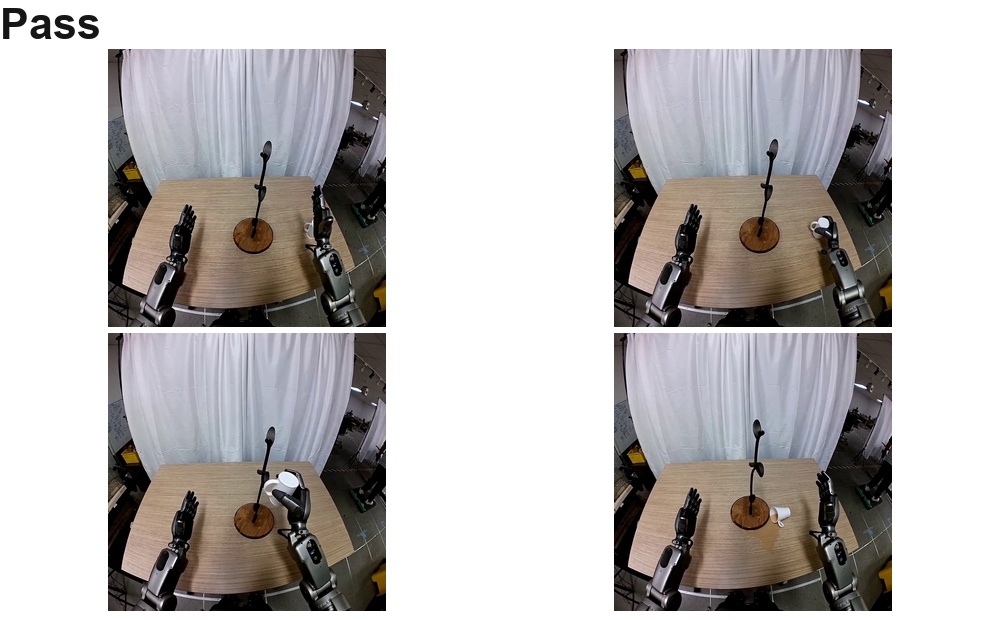}\hfill
\includegraphics[width=\dimexpr(\textwidth-\columnsep)/2\relax]{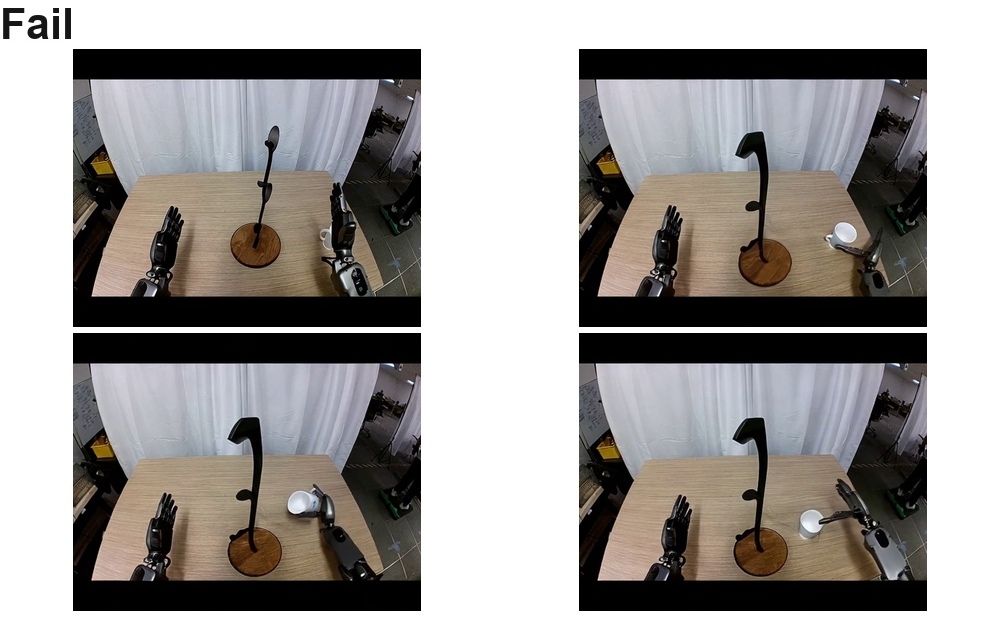}
\captionof{figure}{Human-labeled instruction-following outcomes for the same cup task: lift a white cup, bump the stand, and leave the cup tipped on its side. The left demo leaves the cup on its side; the right leaves it upright. }
\label{fig:appendix_b_preference}
\par\vspace{12pt}
\end{minipage}
}]

\section{Agreement with Human Judgments of Instruction Following}
\label{app:instruction_following}

\paragraph{Evaluation Protocol.}
Instruction following measures whether the video depicts the objects, actions, and outcome specified in the instruction.
Figure~\ref{fig:appendix_b_preference} shows human-labeled examples of instruction-following success and failure for the same task.
We use the same set of 200 videos as in the reward model evaluation in the main paper.
For each model, we pair its instruction-following, semantic-adherence, or text-alignment score, as applicable, with the human IF label for the same video.
For WorldPhyReward, we use its binary IF judgment (R7), which tests whether the video visibly satisfies the instruction's core requirements for objects, actions, relations, event order, and final state.
We compute Pearson and Spearman correlations separately for each run over the available score--label pairs.
Table~\ref{tab:instruction_following_quality} reports the mean of each coefficient across the three runs.

\paragraph{Results.}
WorldPhyReward and RoboJudge-9B show similar agreement with human IF labels: WorldPhyReward has the highest mean Pearson correlation, and RoboJudge-9B has the highest mean Spearman correlation.

\newpage
\begingroup
\centering
\small
\setlength{\tabcolsep}{3pt}
\renewcommand{\arraystretch}{1.1}
\captionsetup{hypcap=false}
\captionof{table}{Mean correlations between model scores and human IF labels. Bold indicates the highest mean in each column.}
\label{tab:instruction_following_quality}
\begin{tabular*}{\columnwidth}{@{\extracolsep{\fill}}lrr@{}}
\toprule
\textbf{Model} & \textbf{Pearson $r$} & \textbf{Spearman $\rho$} \\
\midrule
VideoCon-Physics~\cite{bansal2025videophy}
  & $-0.1082$ & $-0.1124$ \\
VideoPhy-2-AutoEval~\cite{bansal2026videophy}
  & $-0.0359$ & $-0.0340$ \\
VideoScore2~\cite{he2025videoscore2}
  & 0.1220 & 0.1352 \\
WorldModelBench~\cite{li2026worldmodelbench}
  & 0.1383 & 0.1441 \\
VideoReward~\cite{liu2026improving}
  & 0.1818 & 0.1783 \\
UnifiedReward-2.0~\cite{wang2025unified}
  & 0.2011 & 0.1949 \\
RoboJudge-9B~\cite{ma2026robojudge}
  & 0.3445 & \textbf{0.3498} \\
WorldPhyReward (ours)
  & \textbf{0.3469} & 0.3469 \\
\bottomrule
\end{tabular*}

\par\endgroup
\clearpage
\begingroup
\twocolumn[{
\begin{minipage}{\textwidth}
\centering
\footnotesize
\setlength{\tabcolsep}{1.5pt}
\renewcommand{\arraystretch}{1.08}
\captionsetup{type=table}\caption{World-model variants and generation settings. ``Area\textquotesingle\textquotesingle{} denotes target pixel area; actual and bucketed dimensions depend on the input aspect ratio.}
\label{tab:world_model_generation}
\begin{tabular}{@{}>{\raggedright\arraybackslash}p{.251\linewidth}
                       >{\raggedright\arraybackslash}p{.177\linewidth}
                       >{\raggedright\arraybackslash}p{.543\linewidth}@{}}
\toprule
Model / variant & Model size & Generation settings \\
\midrule
Seedance 2.0 & Undisclosed & API model \texttt{doubao-seedance-2-0-260128}; single-reference-image conditioning; 480p request, 1:1, 15\,s; audio and watermark off. \\
Kling 3.0 & Undisclosed & API I2V; 720p request, 15\,s, audio off, multi-shot disabled. \\
MiniMax H3 (FL2VA) & 33B & First-frame-conditioned generation; $960\times768$, 362 frames at 24 fps; 50 steps; video/audio flow shifts 12/3; BF16. \\
Cosmos 3 (Nano) & 16B & First-frame-conditioned generation; $480\times384$, 385 frames at 24 fps; 35 steps, guidance 6, flow shift 10; BF16. \\
HunyuanVideo-1.5 (480P I2V) & 8.3B & Standard, non-distilled checkpoint; 480p buckets, 241 frames at 24 fps; 50 steps, guidance 6; BF16. \\
Wan2.2-I2V-A14B & A14B & I2V; $832\times480$ area, 81 frames at 16 fps; UniPC, 40 steps, guidance $(3.5,3.5)$, shift 3. \\
Wan2.2-TI2V-5B & 5B & Image-conditioned TI2V; $1280\times704$ area, 121 frames at 24 fps; UniPC, 50 steps, guidance 5, shift 5. \\
LongCat-Video & 13.6B & Non-distilled I2V; 480p buckets, 93 frames at 15 fps; 50 steps, guidance 4; BF16. \\
Kairos3.1-4B-robot-480P & 4B & Robot I2V checkpoint; $832\times480$, 81 frames at 16 fps; 50 steps, guidance 5. \\
WoW-1-Wan-14B-2M & 14B & I2V; $832\times480$, 101 frames at 15 fps; 50 steps, guidance 5; BF16. \\
LTX-2.5 Dev & 22B & Two-stage I2V with distilled LoRA in stage 2; $1536\times1024$, 361 frames at 24 fps; 30-step setting; FP8 casting and compilation. \\
LTX-2.5 Distilled & 22B & Distilled I2V pipeline; $1536\times1024$, 361 frames at 24 fps; FP8 casting and internal temporal chunking. \\
CogVideoX1.5-5B-I2V & 5B & I2V; short side 768, long side at most 1360; 161 frames at 16 fps; DPM, 50 steps, dynamic guidance 6; BF16. \\
\bottomrule
\end{tabular}
\par\vspace{12pt}
\end{minipage}
}]

\section{World Model Evaluation}
\label{app:world_model_examples}
\subsection{Generation Setup}
We evaluate 13 world models on a fixed set of 300 tasks, each consisting of a text instruction and an initial image. Each model generates one video per task, except Seedance 2.0, whose API rejected seven inputs flagged as involving human faces, leaving 293 videos for evaluation. This yields 3,893 videos in total. We retain the original instructions and provide no ground-truth future frames or action trajectories. Table~\ref{tab:world_model_generation} lists the generation settings used for each model.

\paragraph{Model variants and inference settings.}
Model sizes follow the released checkpoint designations and exclude auxiliary encoders and autoencoders. I2V denotes image-to-video generation. TI2V supports both text-to-video and image-to-video generation; we use its image-conditioned mode. Seedance 2.0 and Kling 3.0 are accessed through APIs. The 480p and 720p entries for Seedance and Kling specify the requested resolution tiers.

\paragraph{Input processing.}
MiniMax H3 and Cosmos 3 preserve aspect ratio with resizing and padding; Kairos and WoW resize directly to $832\times480$. Wan2.2-I2V-A14B uses aspect-ratio- and area-dependent grid sizes with bicubic resizing; Wan2.2-TI2V-5B uses native size selection and center cropping. HunyuanVideo-1.5 uses the nearest 480p bucket with resize-to-cover and center cropping. LongCat uses 480p buckets and restores output frames to the input dimensions. CogVideoX selects aspect-ratio-dependent dimensions divisible by 16. Both LTX variants resize proportionally and center-crop to $1536\times1024$.

\subsection{World Model Evaluation Examples}
Panels (a)--(k) show examples from world models. Each includes the task, four frames, and WorldPhyReward's seven judgments from the first evaluation run. Reasons and scores are reproduced verbatim, with field names replaced by the dimension abbreviations. The final page discusses the examples and ambiguous judgments.

\newcommand{\wprQ}{\textquotedbl{}}
\newenvironment{wprOutput}
  {\par\smallskip\hrule height .35pt\vskip 4pt
   \fontencoding{T1}\fontsize{8.5}{10}\selectfont\ttfamily\mdseries\raggedright
   \setlength{\parindent}{0pt}\setlength{\parskip}{0pt}\{}
  {\par\}\par}
\newcommand{\wprField}[4]{%
  \par\addvspace{1.8pt}\hangindent=1.3em\hangafter=1\hspace*{.5em}%
  \wprQ#1\wprQ: \{\wprQ reason\wprQ: \wprQ#2\wprQ,
  \mbox{\wprQ score\wprQ: #3}\}#4}

\begin{figure}[tp]
\textbf{(a) Seedance 2.0}\hfill\textbf{7/7}\par
\smallskip
\noindent\parbox[t][3.5\baselineskip][t]{\linewidth}{\raggedright\small Pinch a corner of the folded cloth and lift it with the right hand while the left hand stays still.}\par
\noindent\hfill\includegraphics[width=\linewidth]{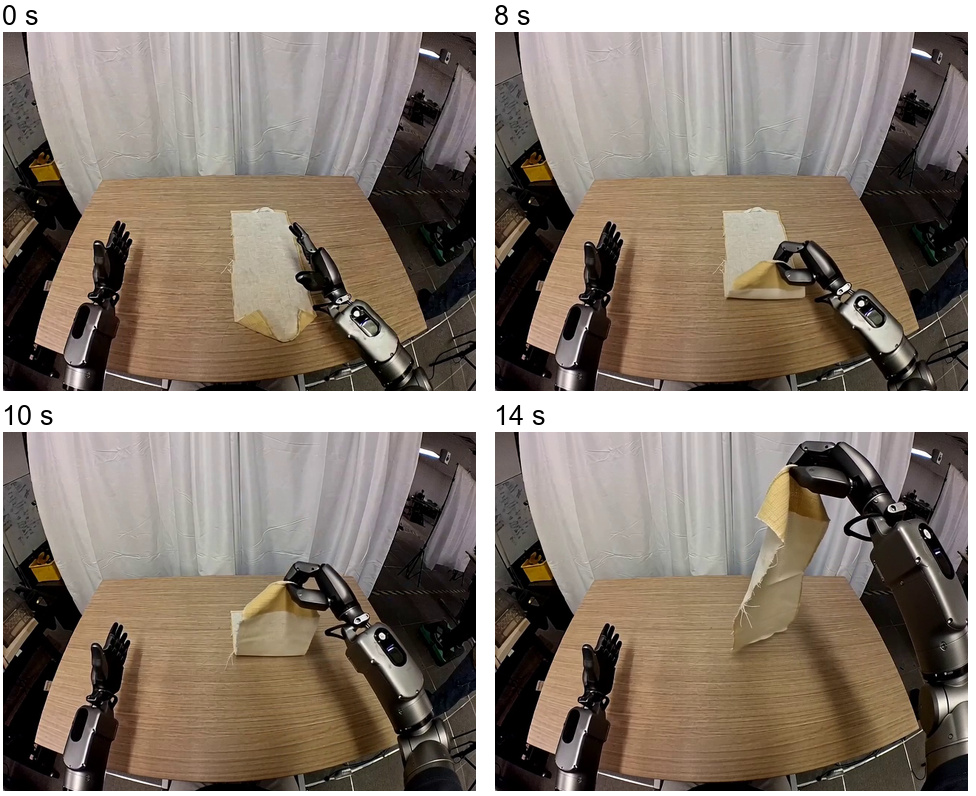}\hfill\null\par
\begin{wprOutput}
\wprField{EIC}{Consistent: Entities remain consistent in existence, count, and identity.}{1}{,}
\wprField{GMP}{Consistent: Shapes and deformations match visible material properties.}{1}{,}
\wprField{CJC}{Consistent: Connections and joint motions respect mechanical constraints.}{1}{,}
\wprField{Penetr.}{Consistent: No visible interpenetration or clipping is observed.}{1}{,}
\wprField{CIC}{Consistent: Object interactions show plausible contact and force transmission.}{1}{,}
\wprField{SMD}{Consistent: Support and motion remain physically plausible.}{1}{,}
\wprField{IF}{Consistent: The video fulfills the caption\textquotesingle{}s core requirements.}{1}{}
\end{wprOutput}
\caption{WorldPhyReward evaluation of Seedance 2.0: sampled frames and seven-dimensional scores with rationales.}
\label{fig:worldphy_example_a}
\end{figure}

\begin{figure}[tp]
\textbf{(b) MiniMax H3}\hfill\textbf{5/7}\par
\smallskip
\noindent\parbox[t][3.5\baselineskip][t]{\linewidth}{\raggedright\small Lift the white cup with the blue lid, carry it toward the back of the table, then lower or release it and return with the gripper empty.}\par
\noindent\hfill\includegraphics[width=\linewidth]{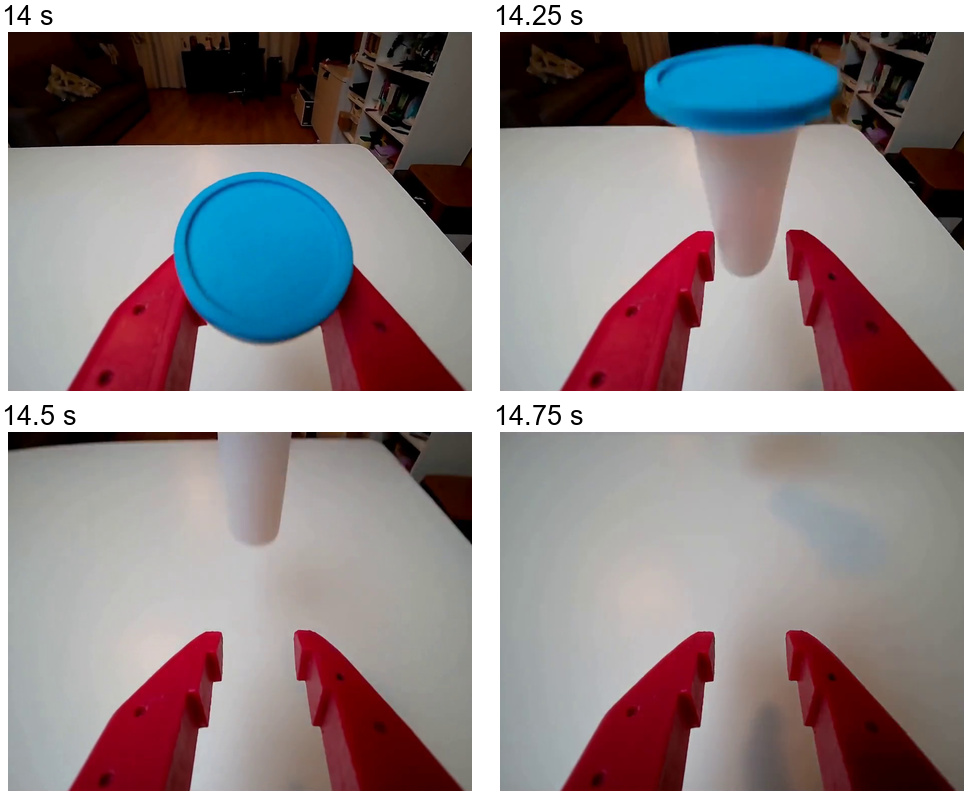}\hfill\null\par
\begin{wprOutput}
\wprField{EIC}{Consistent: Entities remain consistent in existence, count, and identity.}{1}{,}
\wprField{GMP}{Consistent: Shapes and deformations match visible material properties.}{1}{,}
\wprField{CJC}{Consistent: Connections and joint motions respect mechanical constraints.}{1}{,}
\wprField{Penetr.}{Consistent: No visible interpenetration or clipping is observed.}{1}{,}
\wprField{CIC}{Issue: The red gripper tips close beside the cup, yet the cup moves with them without a visible secure grasp.}{0}{,}
\wprField{SMD}{Issue: The cup rises off the tabletop and remains suspended between the open red jaws without visible support.}{0}{,}
\wprField{IF}{Consistent: The video fulfills the caption\textquotesingle{}s core requirements.}{1}{}
\end{wprOutput}
\caption{WorldPhyReward evaluation of MiniMax H3: sampled frames and seven-dimensional scores with rationales.}
\label{fig:worldphy_example_b}
\end{figure}

\begin{figure}[tp]
\textbf{(c) Cosmos 3}\hfill\textbf{2/7}\par
\smallskip
\noindent\parbox[t][3.5\baselineskip][t]{\linewidth}{\raggedright\small Pick up the red apple, place it in the yellow bowl, and retract the right hand.}\par
\noindent\hfill\includegraphics[width=\linewidth]{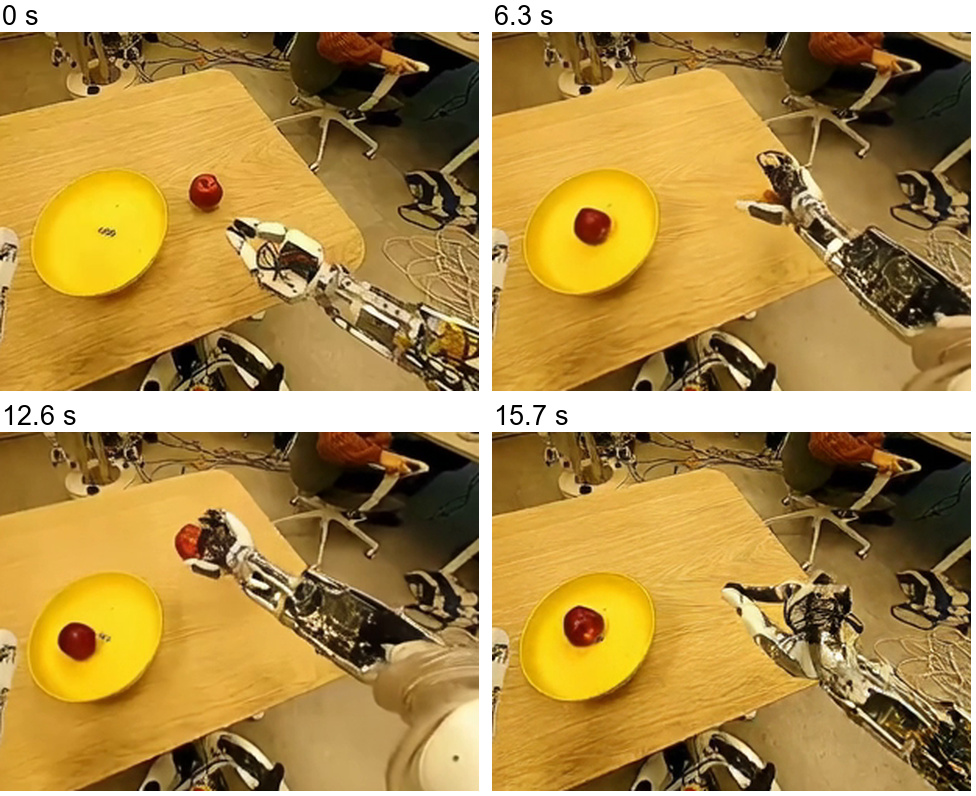}\hfill\null\par
\begin{wprOutput}
\wprField{EIC}{Issue: The red apple duplicates, leaving another apple in the right hand while the first remains in the yellow bowl.}{0}{,}
\wprField{GMP}{Issue: The right robotic hand’s rigid finger segments warp and change thickness during grasping and release.}{0}{,}
\wprField{CJC}{Consistent: Connections and joint motions respect mechanical constraints.}{1}{,}
\wprField{Penetr.}{Issue: The carried apple appears to merge with the apple already in the bowl during release.}{0}{,}
\wprField{CIC}{Issue: The second apple moves with the open right hand without a visible secure grasp.}{0}{,}
\wprField{SMD}{Issue: After release, the apple in the bowl abruptly shifts sideways without a visible impact.}{0}{,}
\wprField{IF}{Consistent: The video fulfills the caption\textquotesingle{}s core requirements.}{1}{}
\end{wprOutput}
\caption{WorldPhyReward evaluation of Cosmos 3: sampled frames and seven-dimensional scores with rationales.}
\label{fig:worldphy_example_c}
\end{figure}

\begin{figure}[tp]
\textbf{(d) Kling 3.0}\hfill\textbf{6/7}\par
\smallskip
\noindent\parbox[t][3.5\baselineskip][t]{\linewidth}{\raggedright\small Pick up the blue bone-shaped toy and release it into the cardboard box.}\par
\noindent\hfill\includegraphics[width=\linewidth]{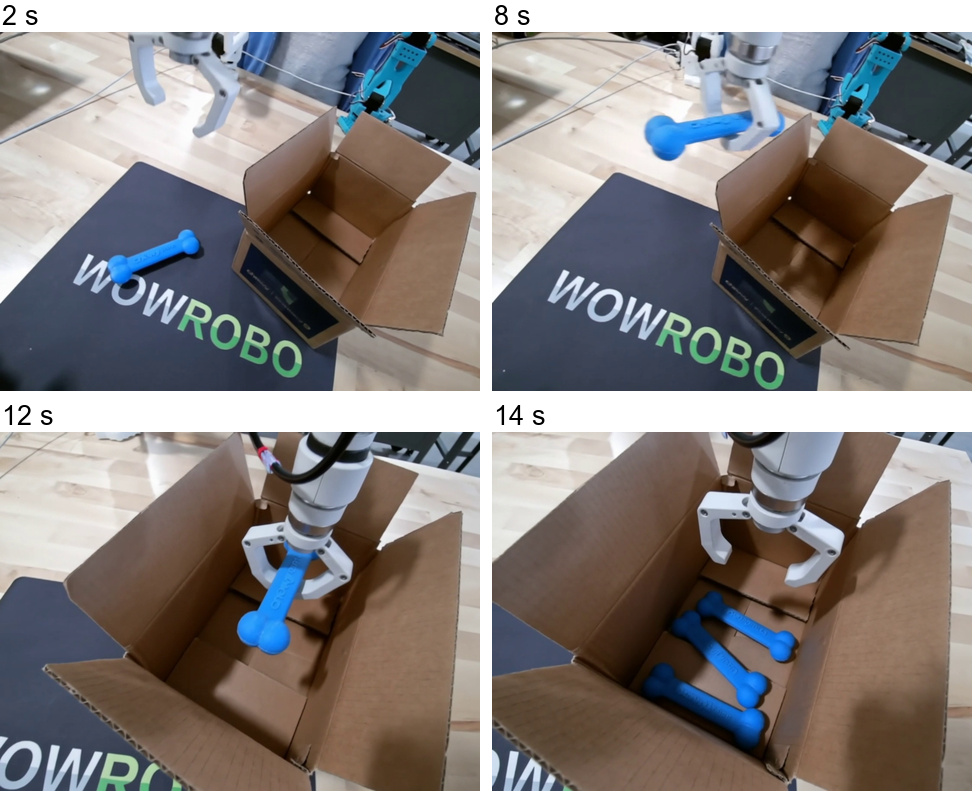}\hfill\null\par
\begin{wprOutput}
\wprField{EIC}{Issue: The blue bone-shaped toy duplicates during release, leaving three toys in the previously empty box.}{0}{,}
\wprField{GMP}{Consistent: Shapes and deformations match visible material properties.}{1}{,}
\wprField{CJC}{Consistent: Connections and joint motions respect mechanical constraints.}{1}{,}
\wprField{Penetr.}{Consistent: No visible interpenetration or clipping is observed.}{1}{,}
\wprField{CIC}{Consistent: Object interactions show plausible contact and force transmission.}{1}{,}
\wprField{SMD}{Consistent: Support and motion remain physically plausible.}{1}{,}
\wprField{IF}{Consistent: The video fulfills the caption\textquotesingle{}s core requirements.}{1}{}
\end{wprOutput}
\caption{WorldPhyReward evaluation of Kling 3.0: sampled frames and seven-dimensional scores with rationales.}
\label{fig:worldphy_example_d}
\end{figure}

\begin{figure}[tp]
\textbf{(e) HunyuanVideo-1.5}\hfill\textbf{4/7}\par
\smallskip
\noindent\parbox[t][3.5\baselineskip][t]{\linewidth}{\raggedright\small Pull open the upper wooden drawer and leave the lower drawer closed.}\par
\noindent\hfill\includegraphics[width=\linewidth]{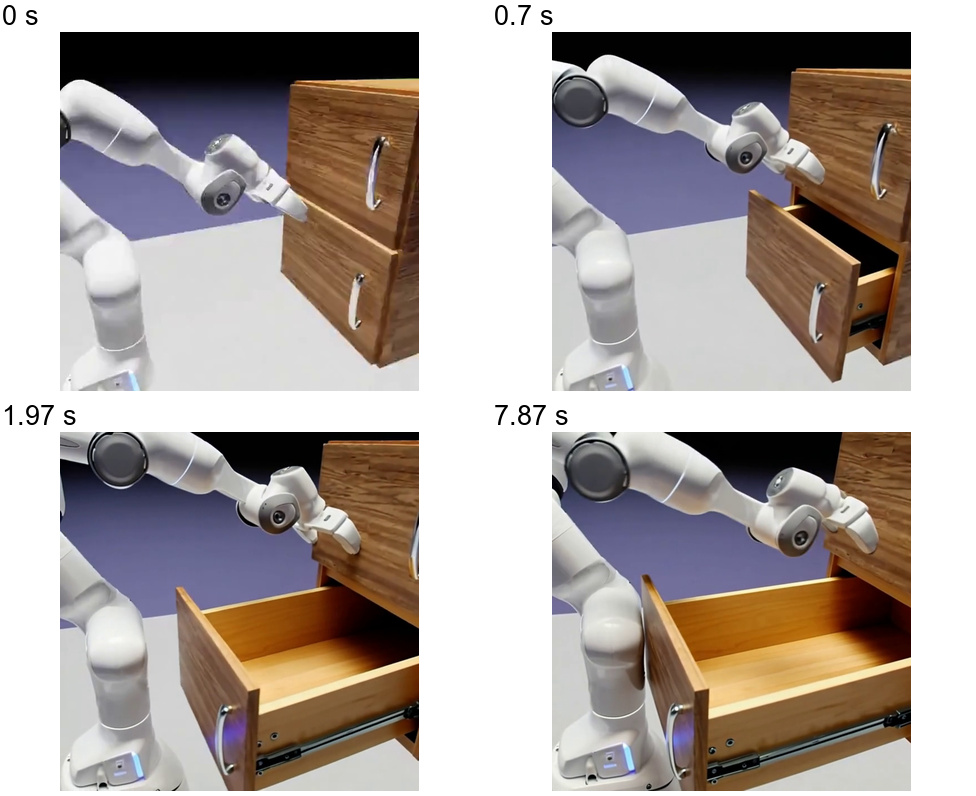}\hfill\null\par
\begin{wprOutput}
\wprField{EIC}{Consistent: Entities remain consistent in existence, count, and identity.}{1}{,}
\wprField{GMP}{Consistent: Shapes and deformations match visible material properties.}{1}{,}
\wprField{CJC}{Consistent: Connections and joint motions respect mechanical constraints.}{1}{,}
\wprField{Penetr.}{Consistent: No visible interpenetration or clipping is observed.}{1}{,}
\wprField{CIC}{Issue: The lower drawer slides open while the gripper remains near the upper drawer, with no visible grasp or pulling contact on the lower drawer.}{0}{,}
\wprField{SMD}{Issue: The lower drawer starts sliding outward from rest without a visible push, pull, or other force initiating its motion.}{0}{,}
\wprField{IF}{Issue: The lower drawer opens and remains open, while the upper drawer stays closed despite the gripper approaching its handle.}{0}{}
\end{wprOutput}
\caption{WorldPhyReward evaluation of HunyuanVideo-1.5: sampled frames and seven-dimensional scores with rationales.}
\label{fig:worldphy_example_e}
\end{figure}

\begin{figure}[tp]
\textbf{(f) Wan2.2-TI2V-5B}\hfill\textbf{6/7}\par
\smallskip
\noindent\parbox[t][3.5\baselineskip][t]{\linewidth}{\raggedright\small Lift the cucumber with the right hand, place it across the yellow plate, and withdraw while the left hand stays still.}\par
\noindent\hfill\includegraphics[width=\linewidth]{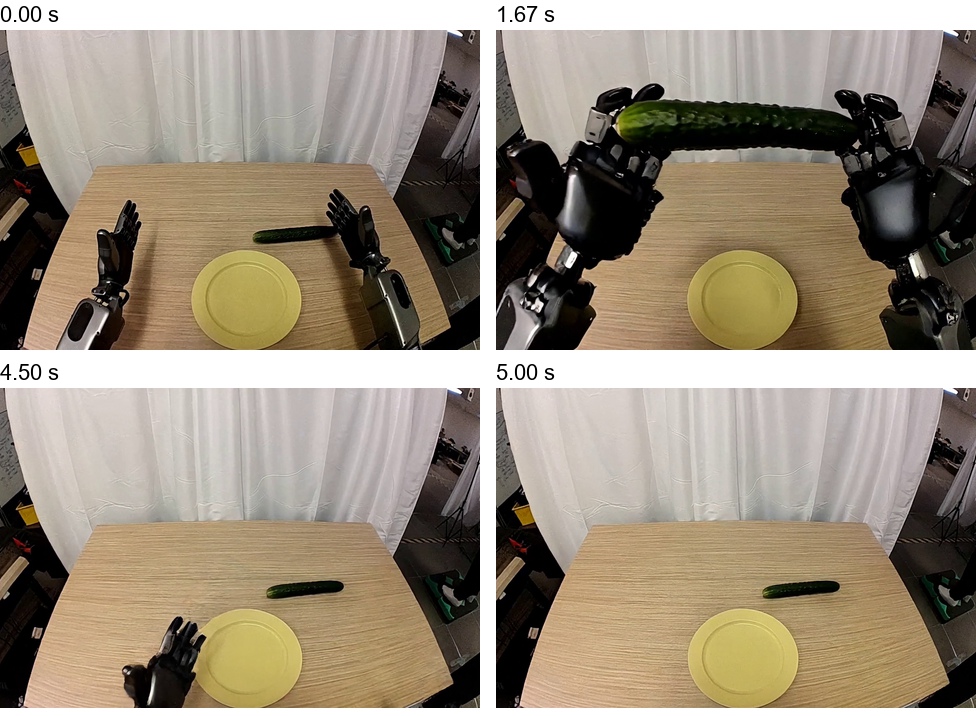}\hfill\null\par
\begin{wprOutput}
\wprField{EIC}{Consistent: Entities remain consistent in existence, count, and identity.}{1}{,}
\wprField{GMP}{Consistent: Shapes and deformations match visible material properties.}{1}{,}
\wprField{CJC}{Consistent: Connections and joint motions respect mechanical constraints.}{1}{,}
\wprField{Penetr.}{Consistent: No visible interpenetration or clipping is observed.}{1}{,}
\wprField{CIC}{Consistent: Object interactions show plausible contact and force transmission.}{1}{,}
\wprField{SMD}{Consistent: Support and motion remain physically plausible.}{1}{,}
\wprField{IF}{Issue: The right hand returns the cucumber to the tabletop beyond the yellow plate, leaving it resting on the table instead of across the plate.}{0}{}
\end{wprOutput}
\caption{WorldPhyReward evaluation of Wan2.2-TI2V-5B: sampled frames and seven-dimensional scores with rationales.}
\label{fig:worldphy_example_f}
\end{figure}

\begin{figure}[tp]
\textbf{(g) Wan2.2-I2V-A14B}\hfill\textbf{3/7}\par
\smallskip
\noindent\parbox[t][3.5\baselineskip][t]{\linewidth}{\raggedright\small Release the blue cube into the white bin and retract the hand.}\par
\noindent\hfill\includegraphics[width=\linewidth]{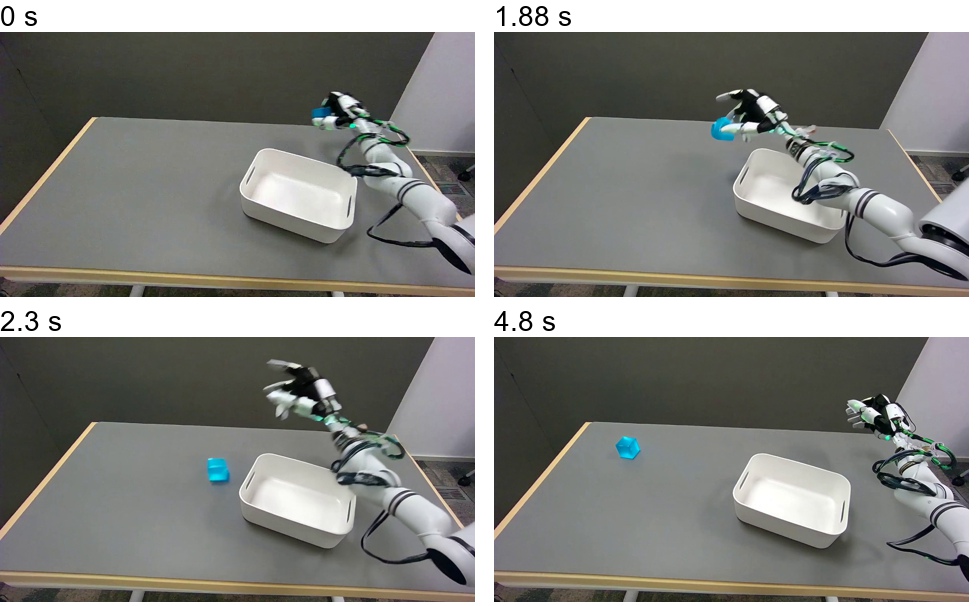}\hfill\null\par
\begin{wprOutput}
\wprField{EIC}{Consistent: Entities remain consistent in existence, count, and identity.}{1}{,}
\wprField{GMP}{Issue: The blue cube briefly becomes a rounded, drooping shape before settling into a cube, making its deformation inconsistent with a rigid object.}{0}{,}
\wprField{CJC}{Consistent: Connections and joint motions respect mechanical constraints.}{1}{,}
\wprField{Penetr.}{Consistent: No visible interpenetration or clipping is observed.}{1}{,}
\wprField{CIC}{Issue: The blue cube separates from the open fingers and moves leftward across the tabletop without a visible push directing it into the bin.}{0}{,}
\wprField{SMD}{Issue: After leaving the hand, the blue cube rises and travels leftward across the tabletop without visible support or propulsion.}{0}{,}
\wprField{IF}{Issue: The blue cube lands on the tabletop beside the bin rather than dropping into it and resting near its left interior wall.}{0}{}
\end{wprOutput}
\caption{WorldPhyReward evaluation of Wan2.2-I2V-A14B: sampled frames and seven-dimensional scores with rationales.}
\label{fig:worldphy_example_g}
\end{figure}

\begin{figure}[tp]
\textbf{(h) LongCat-Video}\hfill\textbf{5/7}\par
\smallskip
\noindent\parbox[t][3.5\baselineskip][t]{\linewidth}{\raggedright\small Pick up the yellow object and release it near the center of the turquoise plate.}\par
\noindent\hfill\includegraphics[width=\linewidth]{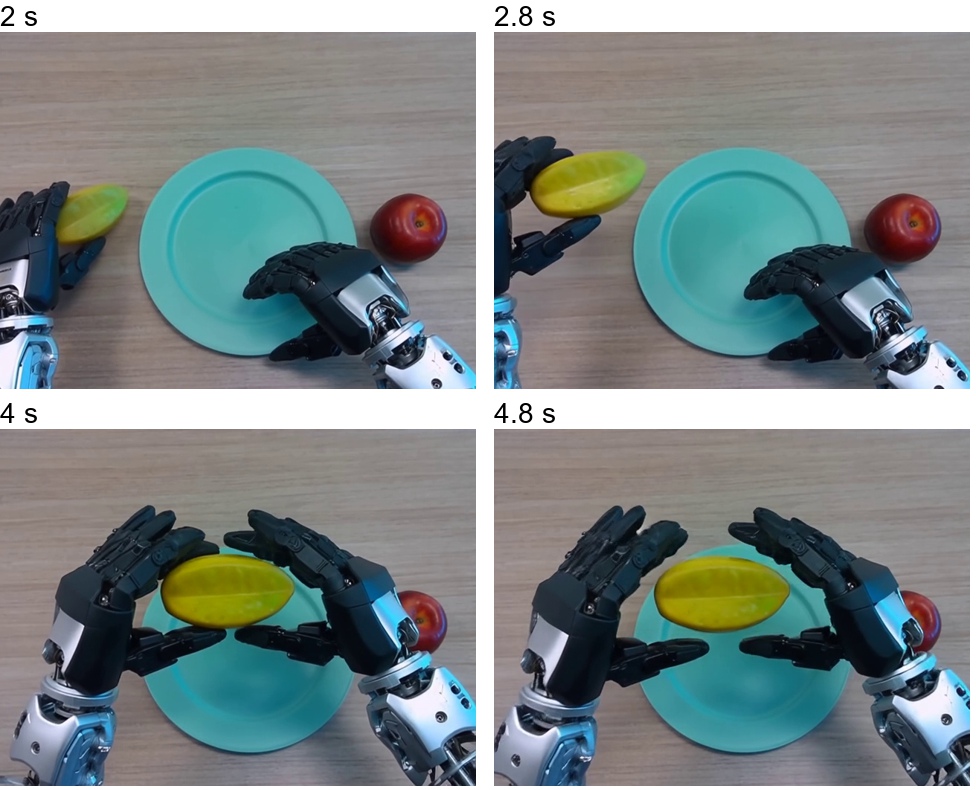}\hfill\null\par
\begin{wprOutput}
\wprField{EIC}{Consistent: Entities remain consistent in existence, count, and identity.}{1}{,}
\wprField{GMP}{Consistent: Shapes and deformations match visible material properties.}{1}{,}
\wprField{CJC}{Consistent: Connections and joint motions respect mechanical constraints.}{1}{,}
\wprField{Penetr.}{Consistent: No visible interpenetration or clipping is observed.}{1}{,}
\wprField{CIC}{Issue: The yellow oval object remains suspended above the plate after both hands withdraw, with no visible contact supporting it.}{0}{,}
\wprField{SMD}{Issue: The yellow oval object hovers above the plate without visible support or downward settling.}{0}{,}
\wprField{IF}{Consistent: The video fulfills the caption\textquotesingle{}s core requirements.}{1}{}
\end{wprOutput}
\caption{WorldPhyReward evaluation of LongCat-Video: sampled frames and seven-dimensional scores with rationales.}
\label{fig:worldphy_example_h}
\end{figure}

\begin{figure}[tp]
\textbf{(i) Kairos3.1-4B}\hfill\textbf{4/7}\par
\smallskip
\noindent\parbox[t][3.5\baselineskip][t]{\linewidth}{\raggedright\small Pour the pieces from the brown cup into the basket, then return the cup upright to the table center.}\par
\noindent\hfill\includegraphics[width=\linewidth]{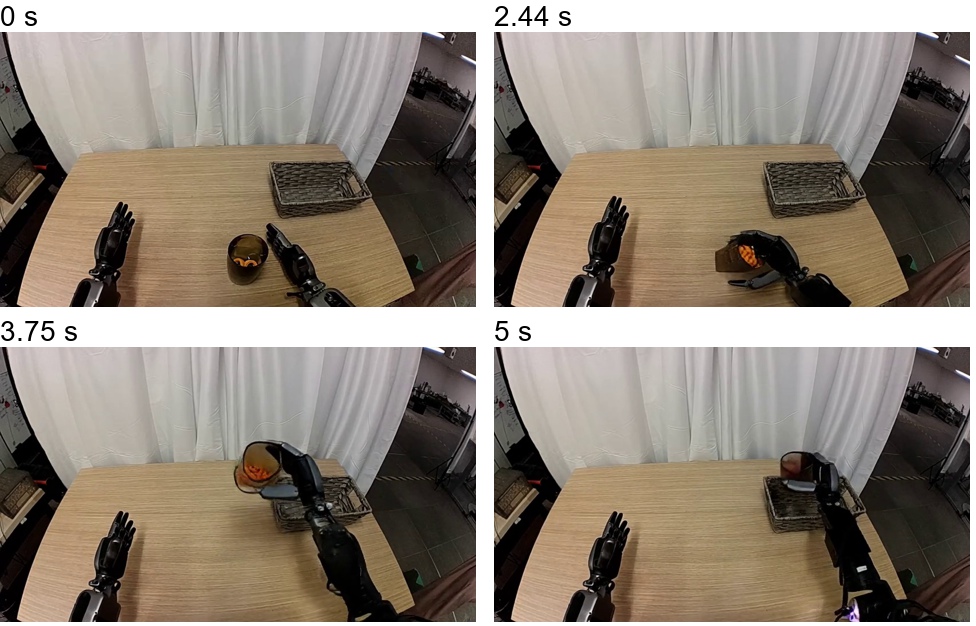}\hfill\null\par
\begin{wprOutput}
\wprField{EIC}{Issue: The orange pieces in the cup multiply during lifting, changing from a few visible pieces into a large pile without a corresponding opening of the cup.}{0}{,}
\wprField{GMP}{Issue: The cup\textquotesingle{}s rim and sidewall distort into an irregular, warped outline as the right hand lifts and carries it.}{0}{,}
\wprField{CJC}{Consistent: Connections and joint motions respect mechanical constraints.}{1}{,}
\wprField{Penetr.}{Consistent: No visible interpenetration or clipping is observed.}{1}{,}
\wprField{CIC}{Consistent: Object interactions show plausible contact and force transmission.}{1}{,}
\wprField{SMD}{Consistent: Support and motion remain physically plausible.}{1}{,}
\wprField{IF}{Issue: The right hand carries the cup over the basket but never visibly pours the orange pieces into it, nor returns the cup upright near the table center.}{0}{}
\end{wprOutput}
\caption{WorldPhyReward evaluation of Kairos3.1-4B: sampled frames and seven-dimensional scores with rationales.}
\label{fig:worldphy_example_i}
\end{figure}

\begin{figure}[tp]
\textbf{(j) WoW-1-Wan-14B}\hfill\textbf{7/7}\par
\smallskip
\noindent\parbox[t][3.5\baselineskip][t]{\linewidth}{\raggedright\small Place the green-handled whisk in the white bowl, release it, and withdraw the hand.}\par
\noindent\hfill\includegraphics[width=\linewidth]{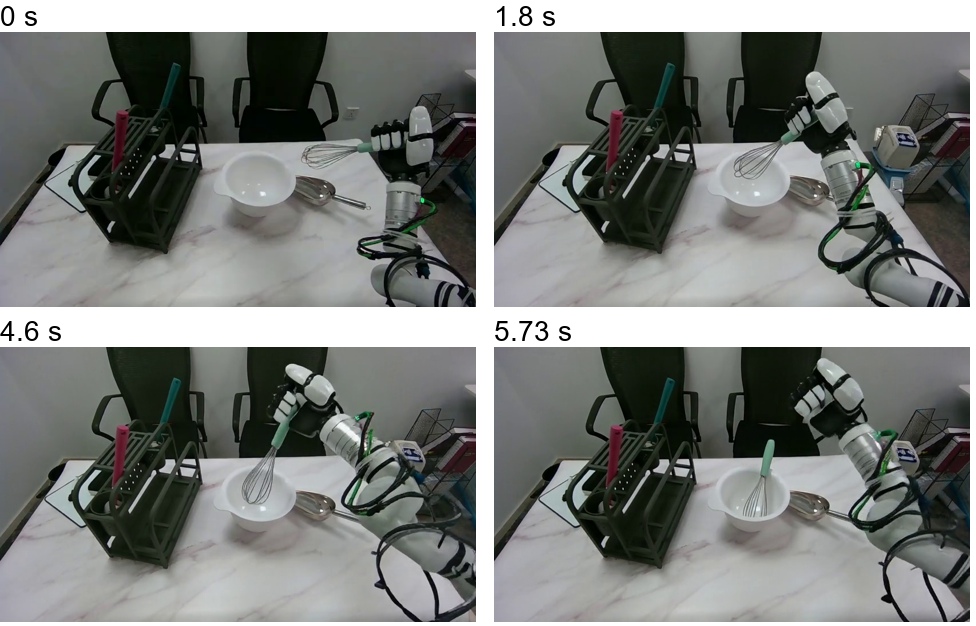}\hfill\null\par
\begin{wprOutput}
\wprField{EIC}{Consistent: Entities remain consistent in existence, count, and identity.}{1}{,}
\wprField{GMP}{Consistent: Shapes and deformations match visible material properties.}{1}{,}
\wprField{CJC}{Consistent: Connections and joint motions respect mechanical constraints.}{1}{,}
\wprField{Penetr.}{Consistent: No visible interpenetration or clipping is observed.}{1}{,}
\wprField{CIC}{Consistent: Object interactions show plausible contact and force transmission.}{1}{,}
\wprField{SMD}{Consistent: Support and motion remain physically plausible.}{1}{,}
\wprField{IF}{Consistent: The video fulfills the caption\textquotesingle{}s core requirements.}{1}{}
\end{wprOutput}
\caption{WorldPhyReward evaluation of WoW-1-Wan-14B: sampled frames and seven-dimensional scores with rationales.}
\label{fig:worldphy_example_j}
\end{figure}

\FloatBarrier
\begin{figure}[tp]
\textbf{(k) LTX-2.5 Distilled}\hfill\textbf{7/7}\par
\smallskip
\noindent\parbox[t][3.5\baselineskip][t]{\linewidth}{\raggedright\small Pick up the drink carton, place it upright on the upper shelf, and withdraw the hand.}\par
\noindent\hfill\includegraphics[width=\linewidth]{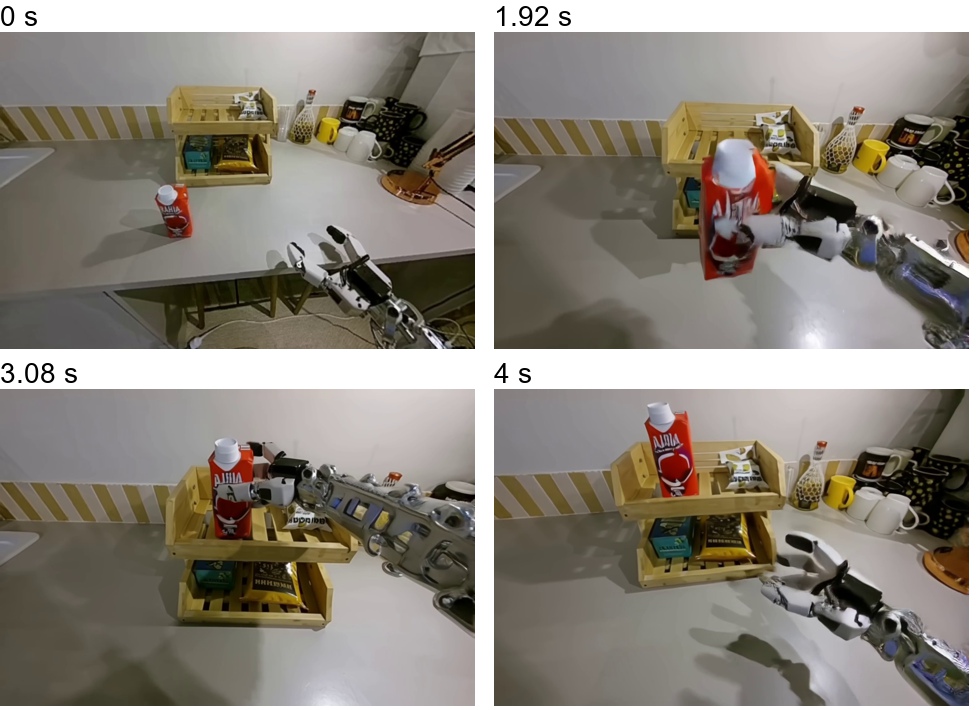}\hfill\null\par
\begin{wprOutput}
\wprField{EIC}{Consistent: Entities remain consistent in existence, count, and identity.}{1}{,}
\wprField{GMP}{Consistent: Shapes and deformations match visible material properties.}{1}{,}
\wprField{CJC}{Consistent: Connections and joint motions respect mechanical constraints.}{1}{,}
\wprField{Penetr.}{Consistent: No visible interpenetration or clipping is observed.}{1}{,}
\wprField{CIC}{Consistent: Object interactions show plausible contact and force transmission.}{1}{,}
\wprField{SMD}{Consistent: Support and motion remain physically plausible.}{1}{,}
\wprField{IF}{Consistent: The video fulfills the caption\textquotesingle{}s core requirements.}{1}{}
\end{wprOutput}
\caption{WorldPhyReward evaluation of LTX-2.5 Distilled: sampled frames and seven-dimensional scores with rationales.}
\label{fig:worldphy_example_k}
\end{figure}

\paragraph{Analysis of the examples.}
In (a), the right hand lifts one corner of the folded cloth while the left hand stays still. In (b), a gap is visible between the cup and the red fingertips at 14.25~s, so there is no visible grasp to explain why the cup moves with the gripper (CIC). By 14.5~s, the cup has risen above both fingertips without visible support (SMD).\par\medskip

\noindent In (c), a second apple appears while the first is still in the bowl (EIC). It moves with the open hand (CIC), and the rigid fingers change shape (GMP). The evaluator reports that the apples appear to merge on release (Penetr.) and one shifts sideways without a visible impact (SMD). In (d), the gripper carries one toy to an empty box. By 14~s, three toys lie inside (EIC).\par\medskip

\noindent In (e), the gripper has no visible grasp on the lower drawer (CIC) as it slides open without a visible push or pull (SMD). The upper drawer stays closed despite the instruction to open it (IF). In (f), the hands lift the cucumber but return it to the tabletop beyond the yellow plate, rather than leaving it on the plate (IF).\par\medskip

\noindent In (g), the blue cube briefly loses its rigid shape (GMP) and lands beside the bin rather than inside it (IF). The evaluator also flags the cube’s upward and leftward motion after release without visible contact or support (CIC, SMD). In (h), the yellow object remains above the plate after both hands withdraw, without visible contact or support (CIC, SMD).\par\medskip

\noindent In (i), the orange pieces visibly increase in number during lifting (EIC), and the cup rim and sidewall deform (GMP). The hand tilts the cup toward the basket but does not return it upright to the table center before the video ends (IF). In (j), the hand lowers the whisk into the bowl, releases it, and withdraws, leaving the whisk supported inside the bowl.\par\medskip

\noindent In (k), the hand places the carton on the upper shelf and withdraws, leaving it upright.\par\medskip
\endgroup
\clearpage
\clearpage
\onecolumn
\begingroup
\begingroup
\centering
\small
\setlength{\tabcolsep}{5pt}
\renewcommand{\arraystretch}{1.14}
\captionsetup{hypcap=false}
\captionof{table}{Video reward models and evaluation settings. Frame counts refer to sampled video frames. Resolution limits and resizing follow each model's native processor.}
\label{tab:reward_inference_settings}
\begin{tabular}{@{}p{.23\linewidth}p{.73\linewidth}@{}}
\toprule
Model / variant & Input processing and inference \\
\midrule
VideoCon-Physics & 32 segment-center frames at $224\times224$; BF16; maximum sequence length 256. Extracts physical and semantic entailment probabilities. \\
VideoPhy-2-AutoEval & 32 segment-center frames at $224\times224$; BF16; greedy decoding, maximum sequence length 256. Produces physical-commonsense and semantic-adherence ratings on a 1--5 scale. \\
VideoReward & Uniform sampling at 2 fps; at most 200,704 pixels per frame, with dimensions aligned to multiples of 28; BF16. Uses the native \texttt{detailed\_special} prompt and normalization for visual quality (VQ), motion quality (MQ), and text alignment (TA). \\
VideoScore2 & 2 fps; native Qwen2.5-VL resizing to multiples of 28; BF16. Sampling with temperature 0.7 and at most 1,024 new tokens; native token-probability-based score extraction for visual quality, text alignment, and physical consistency. \\
WorldModelBench & VILA-EWM-Qwen2-1.5B; 16 uniformly sampled frames, dynamic aspect-ratio processing with 384-pixel vision inputs; BF16. Uses the native generation configuration, with chain-of-thought disabled, to answer one instruction, five physical, and two common-sense questions. \\
UnifiedReward-2.0 & Qwen3-VL-8B variant; 16 uniformly sampled JPEG frames and native resizing; BF16, SDPA, greedy decoding, at most 4,096 new tokens. Outputs alignment, physics, and style ratings on a 1--5 scale. \\
RoboJudge-9B & 4 fps, at most 32 frames and 65,536 pixels per frame; thinking disabled. Sampling with temperature 0.7, top-$p$ 1, and at most 2,048 new tokens. Uses native physical-adherence and instruction-alignment prompts. \\
\bottomrule
\end{tabular}
\par\endgroup
\par\medskip
\section{Reward Model Evaluation}
\label{app:reward_model_evaluation}
\label{app:qualitative_comparisons}
\noindent\begin{minipage}[t]{.485\linewidth}\vspace{0pt}
\subsection{Evaluation Setup}
We compare WorldPhyReward with general-purpose VLMs and existing video reward models on the 200-video test set using human ratings for the seven dimensions (Section~\ref{sec:reward_model_quality}). We also evaluate alignment with human preferences on 50 video pairs from seen generators and 50 from unseen generators (Section~\ref{sec:human_alignment}).

\paragraph{General-purpose VLMs.}
The general-purpose VLM baselines comprise Qwen3.5-27B, Qwen3.6-27B, Qwen3.8-27B, Claude Opus 5.5, Seed2.1 Pro, GPT-5.6 Sol, GPT-6 Astra, and Gemini 3.8 Flash. The three Qwen models run locally, while the other five are accessed through APIs. The evaluation prompt supplies the initial image, chronological video frames sampled at a target rate of 5 fps (up to 128 frames), and the task instruction. It requests seven binary judgments in the order EIC, GMP, CJC, Penetr., CIC, SMD, and IF. Each judgment contains a reason of at most 45 English words, prefixed with \texttt{Issue:} or \texttt{Consistent:}, followed by a score of 0 or 1. WorldPhyReward uses the same output format; its training and inference settings are detailed in Appendix~\ref{app:worldphyreward}.

\end{minipage}\hfill
\begin{minipage}[t]{.485\linewidth}\vspace{0pt}
\paragraph{Video reward models.}
We retain each video reward model's native prompts, video preprocessing, and score extraction. Table~\ref{tab:reward_inference_settings} summarizes the model variants and inference settings. The task instruction is supplied where required by the native evaluator, including its semantic or instruction-alignment component.

\par\csname @minipagefalse\endcsname\csname @nobreakfalse\endcsname
For score correlations, we sum WorldPhyReward's six physical scores for each video. The baselines use their overall physical-plausibility scores, with MQ+VQ for VideoReward and the physical component for WorldModelBench. We compute Pearson and Spearman correlations between each model's scores and the corresponding human physical-score totals. Instruction-following correlations use the corresponding semantic or alignment score (Appendix~\ref{app:instruction_following}).

For pairwise preferences, WorldPhyReward sums all seven scores. VideoCon-Physics, VideoPhy-2-AutoEval, and RoboJudge-9B sum their physical and semantic/instruction scores; UnifiedReward-2.0 and VideoScore2 sum their three component scores. VideoReward uses its normalized overall score, and WorldModelBench uses its 0--10 overall score. The higher-scoring video is preferred; equal scores yield a tie. Native score scales are retained, so their numerical values should not be compared directly across evaluators.

\end{minipage}\par\medskip

\clearpage
\twocolumn
\begingroup\centering
\small\setlength{\tabcolsep}{1.6pt}\renewcommand{\arraystretch}{1.15}
\captionsetup{hypcap=false,aboveskip=0pt}
\captionof{table}{Per-dimension scores for the spoon-manipulation example.}
\label{tab:spoon_example_scores}
\begin{tabular}{@{}lcccccccc@{}}
\toprule
Model & EIC & GMP & CJC & Penetr. & CIC & SMD & IF & Total \\
\midrule
Human & 1 & 1 & 1 & 1 & 0 & 0 & 1 & 5/7 \\
WorldPhyReward & 1 & 1 & 1 & 1 & 0 & 0 & 1 & 5/7 \\
Seed2.1 Pro & 1 & 1 & 1 & 0 & 0 & 1 & 1 & 5/7 \\
GPT-5.6 Sol & 1 & 1 & 1 & 1 & 1 & 1 & 0 & 6/7 \\
GPT-6 Astra & 1 & 1 & 1 & 1 & 1 & 1 & 0 & 6/7 \\
Gemini 3.8 Flash & 1 & 1 & 1 & 0 & 1 & 1 & 1 & 6/7 \\
\bottomrule
\end{tabular}
\par\endgroup
\medskip
\subsection{Reward Model Evaluation Examples}
\paragraph{Comparison with general-purpose VLMs.}
We compare WorldPhyReward with four general-purpose VLMs on the spoon-manipulation video in Figure~\ref{fig:vlm_spoon_frames}. The spoon lifts off the counter and moves with the gripper despite incomplete grasping contact. Table~\ref{tab:spoon_example_scores} reports the human labels and model scores for all seven dimensions.
\par\medskip
\noindent\textit{Task: Slide the white spoon toward the near edge of the counter, then retract the arm.}\par\smallskip
\begin{center}
\includegraphics[width=\linewidth]{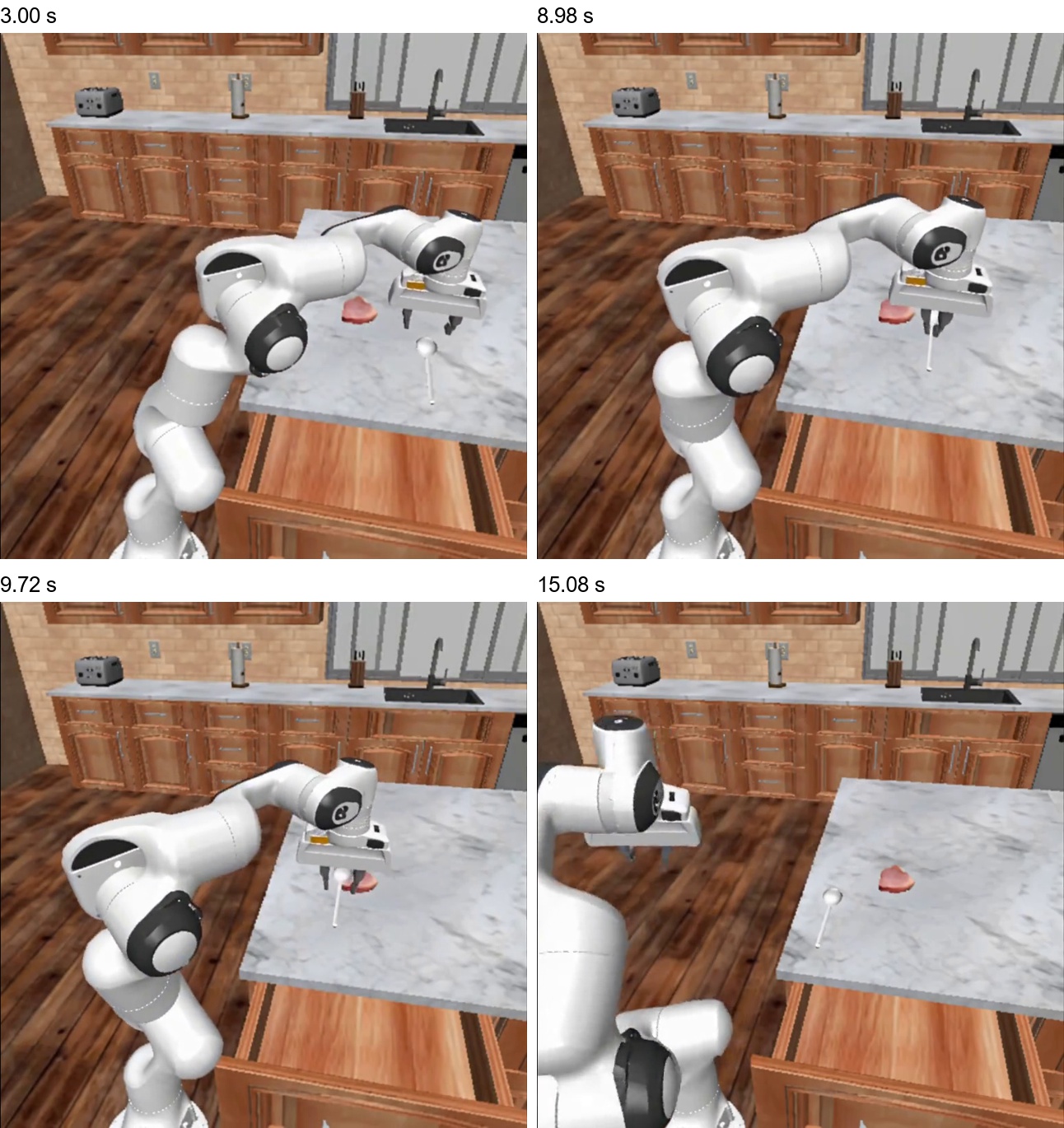}
\begingroup\captionsetup{hypcap=false}
\captionof{figure}{Sampled frames from a spoon-manipulation video generated by MiniMax H3.}
\label{fig:vlm_spoon_frames}\endgroup
\end{center}

\newpage
\noindent The rationales below show how the evaluators judge the contact and support during the spoon's motion. WorldPhyReward flags incomplete grasping contact (CIC) and an unsupported lift (SMD), matching the human labels in all seven dimensions. Seed2.1 Pro also flags CIC. Both Seed2.1 Pro and GPT-5.6 Sol state that the spoon remains supported by the countertop; GPT-6 Astra describes a supported grasp and transport, and Gemini 3.8 Flash describes a contact-driven slide.
\par\medskip
\noindent\begin{minipage}[t]{\linewidth}
\small\textbf{WorldPhyReward}\par\vspace{4pt}\hrule\vspace{5pt}
\noindent CIC (0): Issue: The spoon follows the gripper while one jaw remains visibly separated from it, leaving the apparent pulling contact incomplete.\par\medskip
\noindent SMD (0): Issue: The spoon lifts off the countertop and moves with the gripper despite a visible gap beneath it, making the lift appear unsupported.\par\medskip
\end{minipage}\par\medskip
\noindent\begin{minipage}[t]{\linewidth}
\small\textbf{Seed2.1 Pro}\par\vspace{4pt}\hrule\vspace{5pt}
\noindent CIC (0): Issue: The spoon moves toward the front without a clear sustained grasp or force-transmission contact from the gripper.\par\medskip
\noindent SMD (1): Consistent: The spoon remains supported on the countertop, with no implausible falling, collision, or inertial motion.\par\medskip
\end{minipage}\par\medskip
\noindent\begin{minipage}[t]{\linewidth}
\small\textbf{GPT-5.6 Sol}\par\vspace{4pt}\hrule\vspace{5pt}
\noindent CIC (1): Consistent: The gripper visibly straddles the spoon while moving it, providing a plausible contact path.\par\medskip
\noindent SMD (1): Consistent: The spoon remains counter-supported, moves smoothly with the gripper, and rests after release.\par\medskip
\end{minipage}\par\medskip
\noindent\begin{minipage}[t]{\linewidth}
\small\textbf{GPT-6 Astra}\par\vspace{4pt}\hrule\vspace{5pt}
\noindent CIC (1): Consistent: The gripper closes around the spoon before relocating it and releases it at the destination.\par\medskip
\noindent SMD (1): Consistent: The arm moves smoothly, the spoon remains supported during transport and after placement, and the pink item stays at rest.\par\medskip
\end{minipage}\par\medskip
\noindent\begin{minipage}[t]{\linewidth}
\small\textbf{Gemini 3.8 Flash}\par\vspace{4pt}\hrule\vspace{5pt}
\noindent CIC (1): Consistent: The gripper contacts the spoon and plausibly causes it to slide forward.\par\medskip
\noindent SMD (1): Consistent: The spoon slides along the counter under contact force and halts appropriately.\par\medskip
\end{minipage}\par\medskip

\clearpage
\begingroup\centering
\small\setlength{\tabcolsep}{2pt}\renewcommand{\arraystretch}{1.1}
\captionsetup{hypcap=false,aboveskip=0pt}
\captionof{table}{Scores for the whisk-placement example. Top: human and WorldPhyReward binary scores. Bottom: baseline scores on their native scales.}
\label{tab:whisk_example_scores}
\begin{tabular*}{\linewidth}{@{\extracolsep{\fill}}lcccccccc@{}}
\toprule
Model & EIC & GMP & CJC & Penetr. & CIC & SMD & IF & Total \\
\midrule
Human & 1 & 0 & 1 & 1 & 1 & 1 & 1 & 6/7 \\
WorldPhyReward & 1 & 0 & 1 & 1 & 1 & 1 & 1 & 6/7 \\
\bottomrule
\end{tabular*}\par\smallskip
\begin{tabular*}{\linewidth}{@{\extracolsep{\fill}}p{.30\linewidth}p{.39\linewidth}p{.22\linewidth}@{}}
\toprule
Evaluator & Native score & Value \\
\midrule
WorldModelBench & Physical / Overall & 5/5\enspace /\enspace 10/10 \\
RoboJudge-9B & Physical adherence & 5/5 \\
VideoScore2 & Physical consistency & 3/5 \\
VideoPhy-2-AutoEval & Physical commonsense & 4/5 \\
UnifiedReward-2.0 & Physics & 3.4198/5 \\
VideoReward & Normalized Overall & 0.118774 \\
VideoCon-Physics & Physical adherence probability & 0.041992 \\
\bottomrule
\end{tabular*}
\par\endgroup\smallskip
\medskip
\paragraph{Comparison with video reward models.}
We compare WorldPhyReward with seven existing video reward models on the whisk-placement video in Figure~\ref{fig:reward_whisk_frames}. The fingers flatten as the hand releases the whisk and retracts. Table~\ref{tab:whisk_example_scores} reports human labels and model scores on their native scales.
\par\smallskip

\noindent\textit{Task: Place the green-handled whisk in the white bowl, release it, and retract.}\par\smallskip
\par\smallskip\begingroup\centering
\begin{minipage}{\linewidth}

\begin{minipage}[t]{.484\linewidth}\raggedright\footnotesize 0 s\par\smallskip
\includegraphics[width=\linewidth,trim=96bp 58bp 2016bp 86bp,clip]{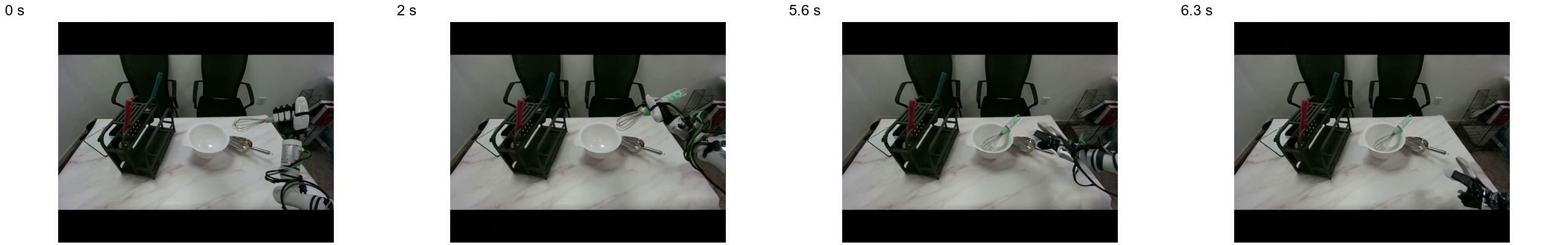}\end{minipage}
\hfill
\begin{minipage}[t]{.484\linewidth}\raggedright\footnotesize 2 s\par\smallskip
\includegraphics[width=\linewidth,trim=736bp 58bp 1376bp 86bp,clip]{fig/appendix_d/cosmos_whisk.jpg}\end{minipage}
\par\smallskip
\begin{minipage}[t]{.484\linewidth}\raggedright\footnotesize 5.6 s\par\smallskip
\includegraphics[width=\linewidth,trim=1376bp 58bp 736bp 86bp,clip]{fig/appendix_d/cosmos_whisk.jpg}\end{minipage}
\hfill
\begin{minipage}[t]{.484\linewidth}\raggedright\footnotesize 6.3 s\par\smallskip
\includegraphics[width=\linewidth,trim=2016bp 58bp 96bp 86bp,clip]{fig/appendix_d/cosmos_whisk.jpg}\end{minipage}

\end{minipage}
\begingroup\captionsetup{hypcap=false}
\captionof{figure}{Sampled frames from a whisk-placement video generated by Cosmos 3.}
\label{fig:reward_whisk_frames}\endgroup
\par\endgroup\smallskip

\par\smallskip

\normalsize
\newpage
\noindent The rationales below show how the evaluators judge the finger deformation. WorldPhyReward identifies the deformation (GMP), matching the human labels in all seven dimensions. WorldModelBench reports no irregular deformation. VideoScore2 misses the finger deformation and instead mentions an abrupt appearance of the bowl. RoboJudge-9B explicitly states that the fingers do not deform. VideoPhy-2-AutoEval, UnifiedReward-2.0, VideoReward, and VideoCon-Physics provide numerical scores without textual rationales.\par\smallskip
\noindent\begin{minipage}[t]{\linewidth}
\small\textbf{WorldPhyReward --- GMP (0)}\par\vspace{3pt}\hrule\vspace{5pt}
Issue: The robot's fingers and palm visibly warp and change thickness as the hand rotates, losing the consistent shape of rigid components.\par\vspace{10pt}
\textbf{WorldModelBench --- original question and answer}\par\vspace{3pt}\hrule\vspace{5pt}
Question: Violation of the Law of Conservation of Mass or Solid Constitutive Law: Objects deform irregularly.\par
Answer: No\par\vspace{10pt}
\textbf{VideoScore2 --- physical-consistency rationale}\par\vspace{3pt}\hrule\vspace{5pt}
Interactions follow plausible physical constraints: the robotic hand grip and motion appear feasible, the whisk is sized reasonably relative to the bowl, and the grasp dynamics are believable. The main caveat is the abrupt appearance of the bowl in the initial frame, which introduces a slight discontinuity in material interaction. Aside from this, gravity and contact behavior are consistent across most frames, supporting a moderate level of physical plausibility (3/5).
\end{minipage}\par\vspace{10pt}
\noindent\begin{minipage}[t]{\linewidth}
\small\textbf{RoboJudge-9B --- complete rationale}\par\vspace{3pt}\hrule\vspace{5pt}
Agent integrity: the white robot arm, black gripper, cables, and green joint stay rigid and connected; no fingers deform or duplicate. Scene \& object consistency: the white marble table, white bowl, metal spoon, utensil rack, chairs, wall, and cables keep stable positions. The bowl remains round and centered; the spoon and rack do not shift or resize. Interaction realism: the arm reaches from the right, carries the silver wire whisk, and lowers it into the bowl in a plausible path. Contact is mostly believable; the whisk enters the bowl, settles inside, and stays supported by gravity. Minor issues include slightly hidden grasp contact and a little abrupt settling, but no clear physical violation.\par\smallskip
\end{minipage}\par
\par\smallskip

\par

\twocolumn[{
\noindent
\begin{minipage}{\textwidth}

\begin{minipage}[t]{.47\linewidth}\vspace{0pt}
\noindent\textbf{(a) Stacking}\par
\noindent\textit{Task: Stack the blue piece on the orange piece, then place the green piece on top.}\par\smallskip
\noindent\hfill\begin{minipage}[t]{.94\linewidth}\centering\small A --- Seedance 2.0\par\smallskip
\begingroup\setlength{\parskip}{0pt}\noindent
\begin{minipage}[t]{.493\linewidth}\raggedright\scriptsize 0 s\par\vspace{1pt}
\includegraphics[width=\linewidth,trim=136bp 465bp 784bp 95bp,clip]{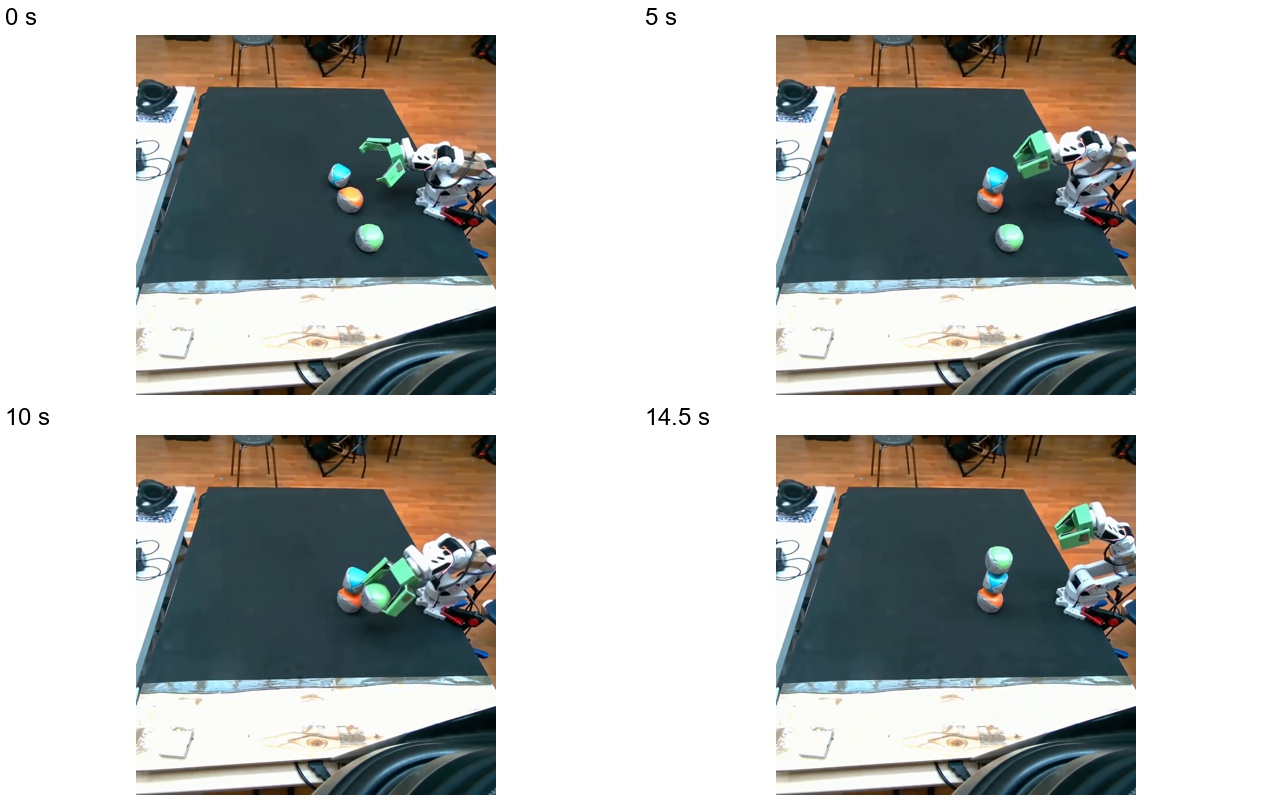}\end{minipage}
\hfill%
\begin{minipage}[t]{.493\linewidth}\raggedright\scriptsize 5 s\par\vspace{1pt}
\includegraphics[width=\linewidth,trim=776bp 465bp 144bp 95bp,clip]{fig/appendix_d/seen_stacking_seedance.jpg}\end{minipage}
\par\vspace{3pt}\noindent
\begin{minipage}[t]{.493\linewidth}\raggedright\scriptsize 10 s\par\vspace{1pt}
\includegraphics[width=\linewidth,trim=136bp 65bp 784bp 495bp,clip]{fig/appendix_d/seen_stacking_seedance.jpg}\end{minipage}
\hfill%
\begin{minipage}[t]{.493\linewidth}\raggedright\scriptsize 14.5 s\par\vspace{1pt}
\includegraphics[width=\linewidth,trim=776bp 65bp 144bp 495bp,clip]{fig/appendix_d/seen_stacking_seedance.jpg}\end{minipage}
\par\endgroup\end{minipage}\hfill\null\par\smallskip
\noindent\hfill\begin{minipage}[t]{.94\linewidth}\centering\small B --- MiniMax H3\par\smallskip
\begingroup\setlength{\parskip}{0pt}\noindent
\begin{minipage}[t]{.493\linewidth}\raggedright\scriptsize 0 s\par\vspace{1pt}
\includegraphics[width=\linewidth,trim=91bp 435bp 739bp 65bp,clip]{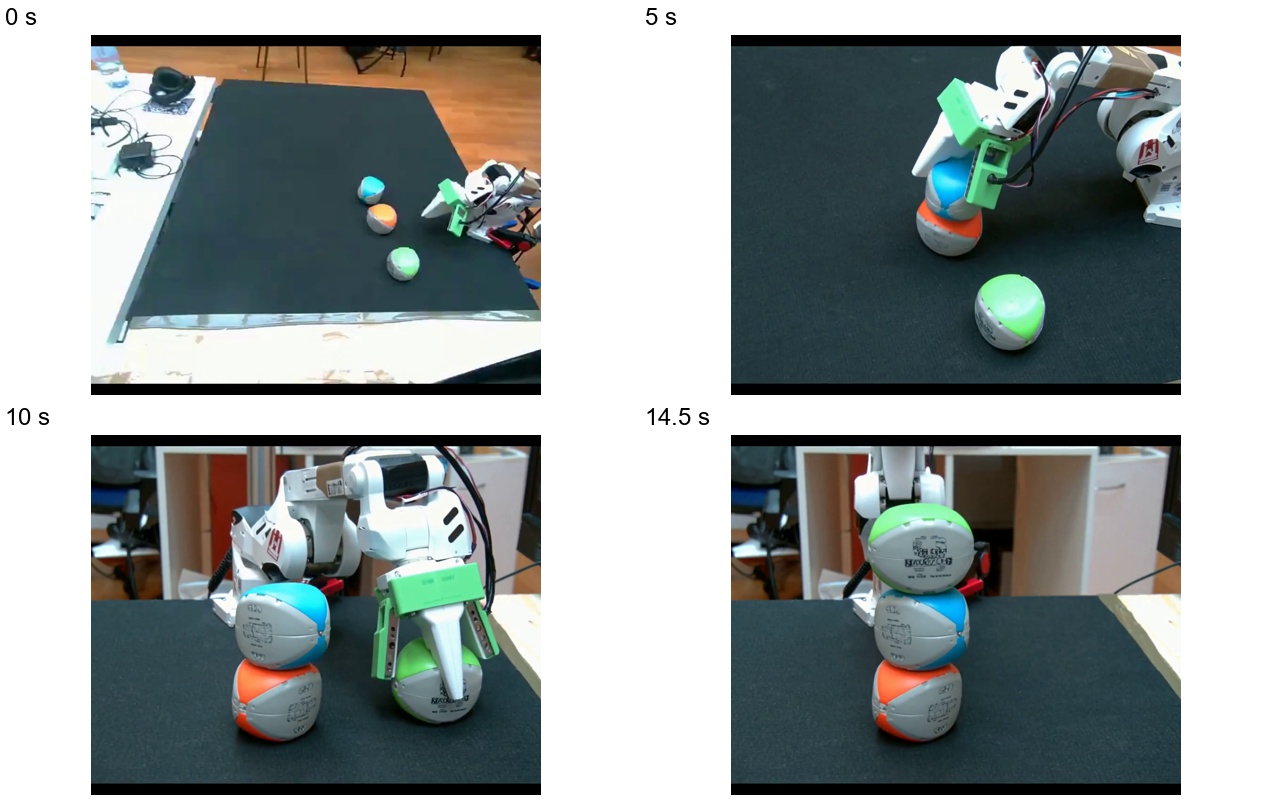}\end{minipage}
\hfill%
\begin{minipage}[t]{.493\linewidth}\raggedright\scriptsize 5 s\par\vspace{1pt}
\includegraphics[width=\linewidth,trim=731bp 435bp 99bp 65bp,clip]{fig/appendix_d/seen_stacking_minimax.jpg}\end{minipage}
\par\vspace{3pt}\noindent
\begin{minipage}[t]{.493\linewidth}\raggedright\scriptsize 10 s\par\vspace{1pt}
\includegraphics[width=\linewidth,trim=91bp 35bp 739bp 465bp,clip]{fig/appendix_d/seen_stacking_minimax.jpg}\end{minipage}
\hfill%
\begin{minipage}[t]{.493\linewidth}\raggedright\scriptsize 14.5 s\par\vspace{1pt}
\includegraphics[width=\linewidth,trim=731bp 35bp 99bp 465bp,clip]{fig/appendix_d/seen_stacking_minimax.jpg}\end{minipage}
\par\endgroup\end{minipage}\hfill\null\par\smallskip
\end{minipage}\hfill
\begin{minipage}[t]{.47\linewidth}\vspace{0pt}
\noindent\textbf{(b) Object placement}\par
\noindent\textit{Task: Place the yellow object upright inside the white bin and release it.}\par\smallskip
\noindent\hfill\begin{minipage}[t]{.94\linewidth}\centering\small A --- Seedance 2.0\par\smallskip
\begingroup\setlength{\parskip}{0pt}\noindent
\begin{minipage}[t]{.493\linewidth}\raggedright\scriptsize 0 s\par\vspace{1pt}
\includegraphics[width=\linewidth,trim=136bp 465bp 784bp 95bp,clip]{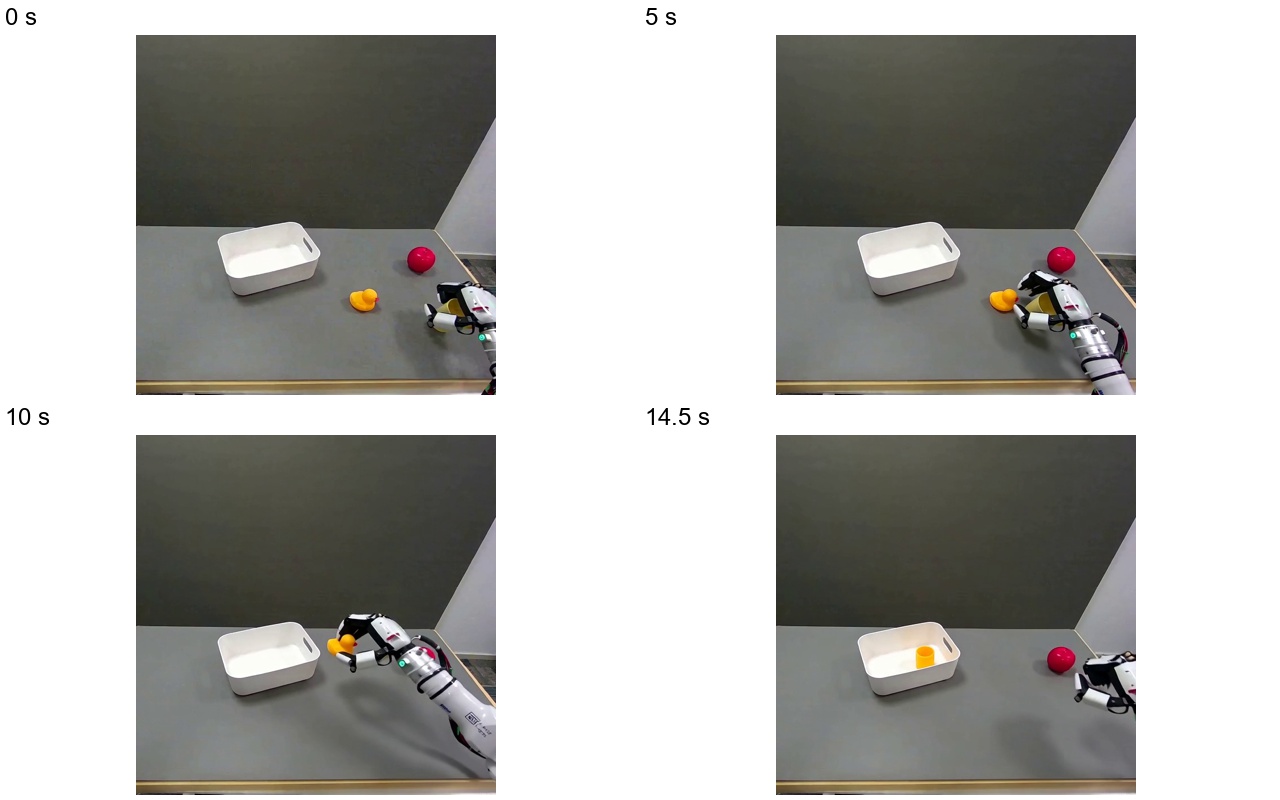}\end{minipage}
\hfill%
\begin{minipage}[t]{.493\linewidth}\raggedright\scriptsize 5 s\par\vspace{1pt}
\includegraphics[width=\linewidth,trim=776bp 465bp 144bp 95bp,clip]{fig/appendix_d/seen_object_placement_seedance.jpg}\end{minipage}
\par\vspace{3pt}\noindent
\begin{minipage}[t]{.493\linewidth}\raggedright\scriptsize 10 s\par\vspace{1pt}
\includegraphics[width=\linewidth,trim=136bp 65bp 784bp 495bp,clip]{fig/appendix_d/seen_object_placement_seedance.jpg}\end{minipage}
\hfill%
\begin{minipage}[t]{.493\linewidth}\raggedright\scriptsize 14.5 s\par\vspace{1pt}
\includegraphics[width=\linewidth,trim=776bp 65bp 144bp 495bp,clip]{fig/appendix_d/seen_object_placement_seedance.jpg}\end{minipage}
\par\endgroup\end{minipage}\hfill\null\par\smallskip
\noindent\hfill\begin{minipage}[t]{.94\linewidth}\centering\small B --- MiniMax H3\par\smallskip
\begingroup\setlength{\parskip}{0pt}\noindent
\begin{minipage}[t]{.493\linewidth}\raggedright\scriptsize 0 s\par\vspace{1pt}
\includegraphics[width=\linewidth,trim=91bp 435bp 739bp 65bp,clip]{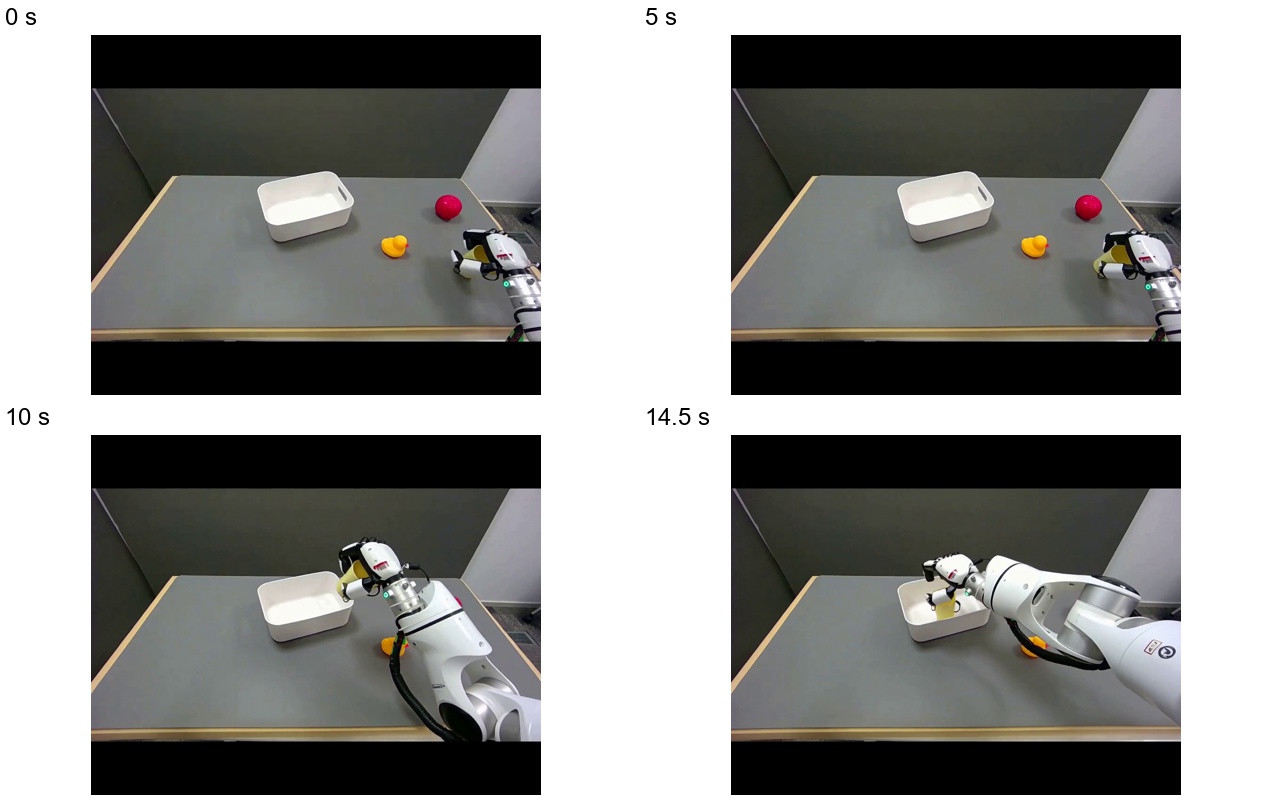}\end{minipage}
\hfill%
\begin{minipage}[t]{.493\linewidth}\raggedright\scriptsize 5 s\par\vspace{1pt}
\includegraphics[width=\linewidth,trim=731bp 435bp 99bp 65bp,clip]{fig/appendix_d/seen_object_placement_minimax.jpg}\end{minipage}
\par\vspace{3pt}\noindent
\begin{minipage}[t]{.493\linewidth}\raggedright\scriptsize 10 s\par\vspace{1pt}
\includegraphics[width=\linewidth,trim=91bp 35bp 739bp 465bp,clip]{fig/appendix_d/seen_object_placement_minimax.jpg}\end{minipage}
\hfill%
\begin{minipage}[t]{.493\linewidth}\raggedright\scriptsize 14.5 s\par\vspace{1pt}
\includegraphics[width=\linewidth,trim=731bp 35bp 99bp 465bp,clip]{fig/appendix_d/seen_object_placement_minimax.jpg}\end{minipage}
\par\endgroup\end{minipage}\hfill\null\par\smallskip
\end{minipage}\par
\begingroup\captionsetup{hypcap=false}
\captionof{figure}{Sampled frames from two seen-generator video pairs: (a) stacking and (b) object placement.}
\label{fig:seen_preference_frames}\endgroup
\end{minipage}
\par\vspace{12pt}
}]
\paragraph{Human preference alignment on seen generators.}
We show two video pairs from the seen-generator evaluation (Figure~\ref{fig:seen_preference_frames}): stacking in (a) and object placement in (b). Table~\ref{tab:seen_example_preferences} compares human preferences with predictions from all reward models in that evaluation.\par
\medskip
\normalsize
\noindent Human annotators express no preference in (a) and prefer B in (b). WorldPhyReward matches both judgments. Among the baselines, RoboJudge-9B matches the tie in (a), while VideoCon-Physics and UnifiedReward-2.0 match the preference for B in (b); none matches both.\par
\newpage
\begingroup
\centering\small
\setlength{\tabcolsep}{2pt}\renewcommand{\arraystretch}{1.15}
\captionsetup{hypcap=false,aboveskip=0pt}
\captionof{table}{Human preference alignment on seen generators. A and B denote the preferred video; Tie denotes no preference.}
\label{tab:seen_example_preferences}
\begin{tabular*}{\columnwidth}{@{\extracolsep{\fill}}lcc@{}}
\toprule
Evaluator & (a) Stacking & \shortstack{(b) Object\\placement} \\
\midrule
Human & Tie & B \\
\midrule
VideoCon-Physics & B & B \\
UnifiedReward-2.0 & A & B \\
VideoPhy-2-AutoEval & A & Tie \\
VideoReward & B & A \\

VideoScore2 & A & A \\
RoboJudge-9B & Tie & Tie \\
WorldModelBench & A & Tie \\
\midrule
WorldPhyReward & Tie & B \\
\bottomrule
\end{tabular*}
\par\endgroup

\twocolumn[{
\noindent
\begin{minipage}{\textwidth}

\begin{minipage}[t]{.47\linewidth}\vspace{0pt}
\noindent\textbf{(c) Apple placement}\par
\noindent\textit{Task: Grasp the red apple, place it on the yellow plate, release it, and retract.}\par\smallskip
\noindent\hfill\begin{minipage}[t]{.94\linewidth}\centering\small A --- Kling 3.0\par\smallskip
\begingroup\setlength{\parskip}{0pt}\noindent
\begin{minipage}[t]{.493\linewidth}\raggedright\scriptsize 0 s\par\vspace{1pt}
\includegraphics[width=\linewidth,trim=79bp 424.333bp 719bp 54.3333bp,clip]{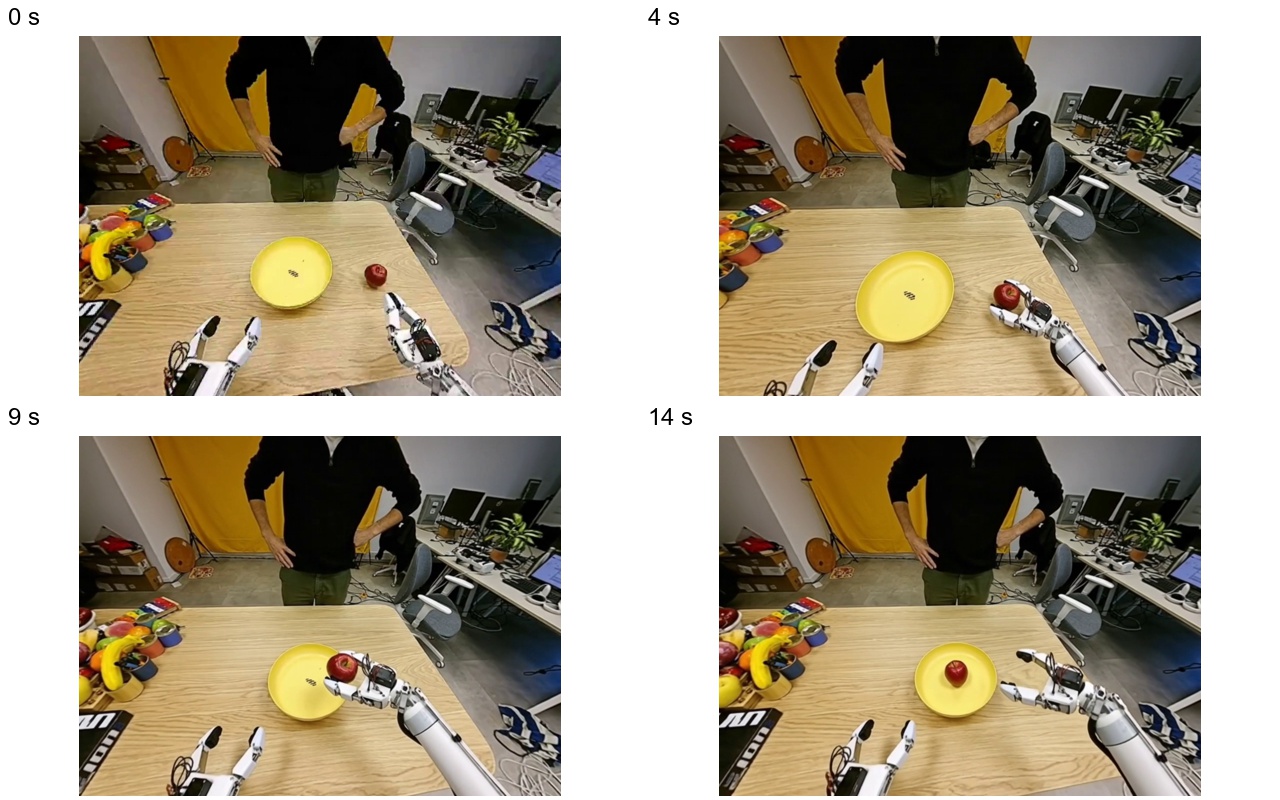}\end{minipage}
\hfill%
\begin{minipage}[t]{.493\linewidth}\raggedright\scriptsize 4 s\par\vspace{1pt}
\includegraphics[width=\linewidth,trim=719bp 424.333bp 79bp 54.3333bp,clip]{fig/appendix_d/unseen_apple_kling.jpg}\end{minipage}
\par\vspace{3pt}\noindent
\begin{minipage}[t]{.493\linewidth}\raggedright\scriptsize 9 s\par\vspace{1pt}
\includegraphics[width=\linewidth,trim=79bp 24.3333bp 719bp 454.333bp,clip]{fig/appendix_d/unseen_apple_kling.jpg}\end{minipage}
\hfill%
\begin{minipage}[t]{.493\linewidth}\raggedright\scriptsize 14 s\par\vspace{1pt}
\includegraphics[width=\linewidth,trim=719bp 24.3333bp 79bp 454.333bp,clip]{fig/appendix_d/unseen_apple_kling.jpg}\end{minipage}
\par\endgroup\end{minipage}\hfill\null\par\smallskip
\hfill\begin{minipage}[t]{.94\linewidth}\centering\small B --- LTX-2.5 Distilled\par\smallskip
\begingroup\setlength{\parskip}{0pt}\noindent
\begin{minipage}[t]{.493\linewidth}\raggedright\scriptsize 0 s\par\vspace{1pt}
\includegraphics[width=\linewidth,trim=50bp 405bp 690bp 35bp,clip]{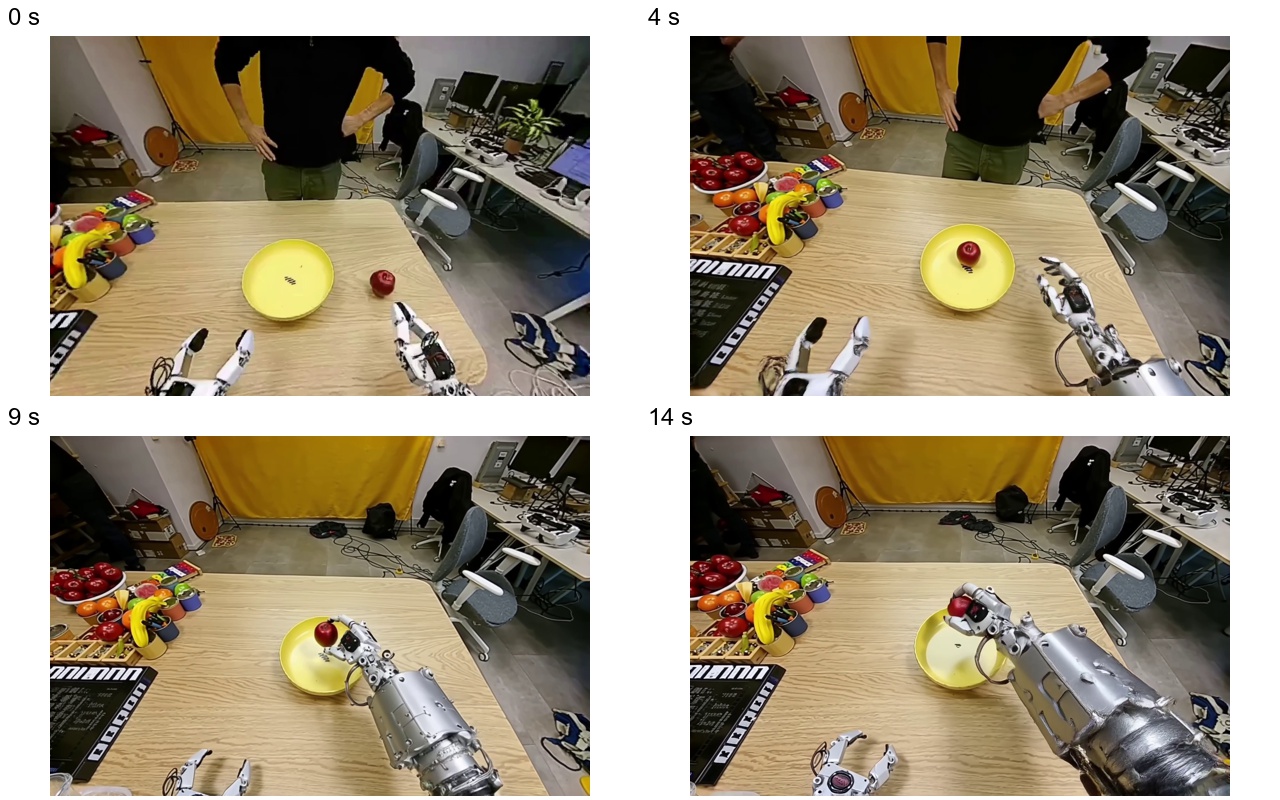}\end{minipage}
\hfill%
\begin{minipage}[t]{.493\linewidth}\raggedright\scriptsize 4 s\par\vspace{1pt}
\includegraphics[width=\linewidth,trim=690bp 405bp 50bp 35bp,clip]{fig/appendix_d/unseen_apple_ltx_distilled.jpg}\end{minipage}
\par\vspace{3pt}\noindent
\begin{minipage}[t]{.493\linewidth}\raggedright\scriptsize 9 s\par\vspace{1pt}
\includegraphics[width=\linewidth,trim=50bp 5bp 690bp 435bp,clip]{fig/appendix_d/unseen_apple_ltx_distilled.jpg}\end{minipage}
\hfill%
\begin{minipage}[t]{.493\linewidth}\raggedright\scriptsize 14 s\par\vspace{1pt}
\includegraphics[width=\linewidth,trim=690bp 5bp 50bp 435bp,clip]{fig/appendix_d/unseen_apple_ltx_distilled.jpg}\end{minipage}
\par\endgroup\end{minipage}\hfill\null\par\smallskip
\end{minipage}\hfill
\begin{minipage}[t]{.47\linewidth}\vspace{0pt}
\noindent\textbf{(d) Apple placement, wrist view}\par
\noindent\textit{Task: Grasp the apple, place it in the white bowl, release it, and withdraw.}\par\smallskip
\noindent\hfill\begin{minipage}[t]{.94\linewidth}\centering\small A --- Kling 3.0\par\smallskip
\begingroup\setlength{\parskip}{0pt}\noindent
\begin{minipage}[t]{.493\linewidth}\raggedright\scriptsize 0 s\par\vspace{1pt}
\includegraphics[width=\linewidth,trim=79bp 424.333bp 719bp 54.3333bp,clip]{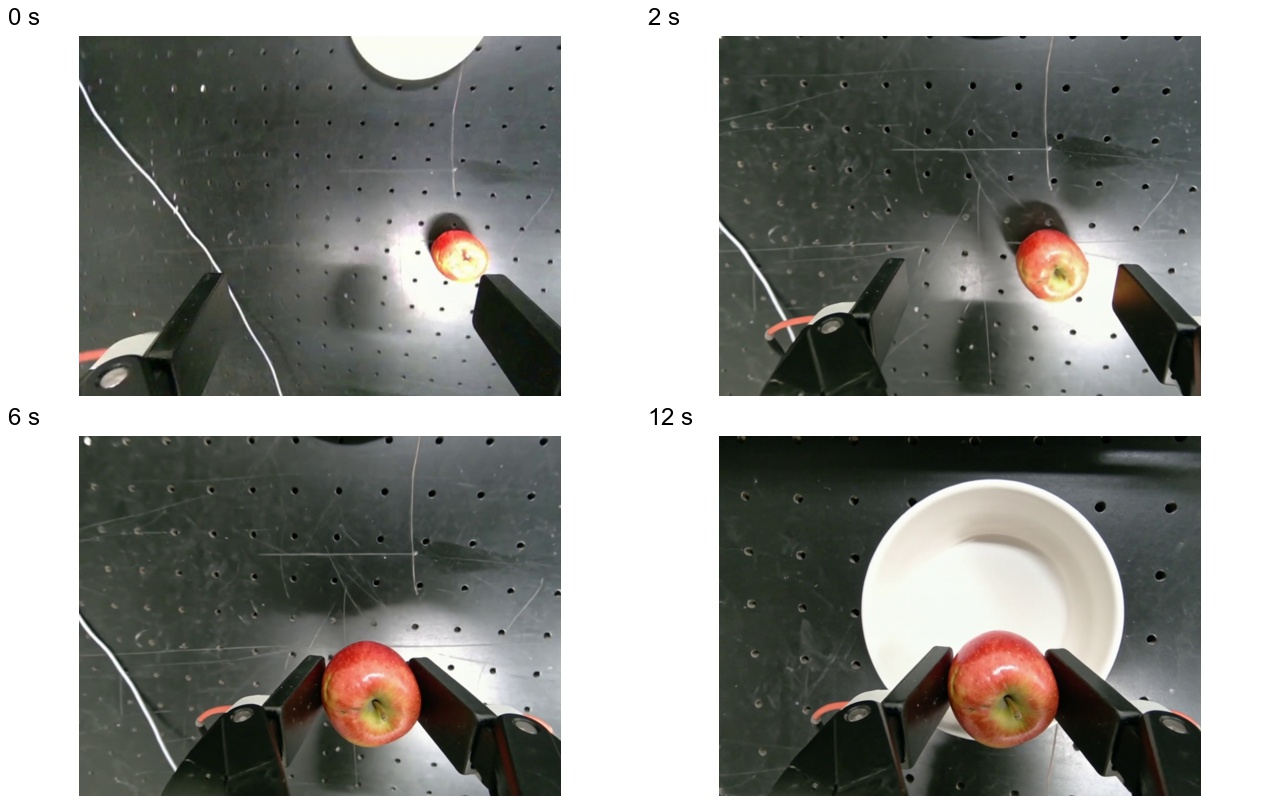}\end{minipage}
\hfill%
\begin{minipage}[t]{.493\linewidth}\raggedright\scriptsize 2 s\par\vspace{1pt}
\includegraphics[width=\linewidth,trim=719bp 424.333bp 79bp 54.3333bp,clip]{fig/appendix_d/unseen_wrist_apple_kling.jpg}\end{minipage}
\par\vspace{3pt}\noindent
\begin{minipage}[t]{.493\linewidth}\raggedright\scriptsize 6 s\par\vspace{1pt}
\includegraphics[width=\linewidth,trim=79bp 24.3333bp 719bp 454.333bp,clip]{fig/appendix_d/unseen_wrist_apple_kling.jpg}\end{minipage}
\hfill%
\begin{minipage}[t]{.493\linewidth}\raggedright\scriptsize 12 s\par\vspace{1pt}
\includegraphics[width=\linewidth,trim=719bp 24.3333bp 79bp 454.333bp,clip]{fig/appendix_d/unseen_wrist_apple_kling.jpg}\end{minipage}
\par\endgroup\end{minipage}\hfill\null\par\smallskip
\hfill\begin{minipage}[t]{.94\linewidth}\centering\small B --- LTX-2.5 Distilled\par\smallskip
\begingroup\setlength{\parskip}{0pt}\noindent
\begin{minipage}[t]{.493\linewidth}\raggedright\scriptsize 0 s\par\vspace{1pt}
\includegraphics[width=\linewidth,trim=50bp 405bp 690bp 35bp,clip]{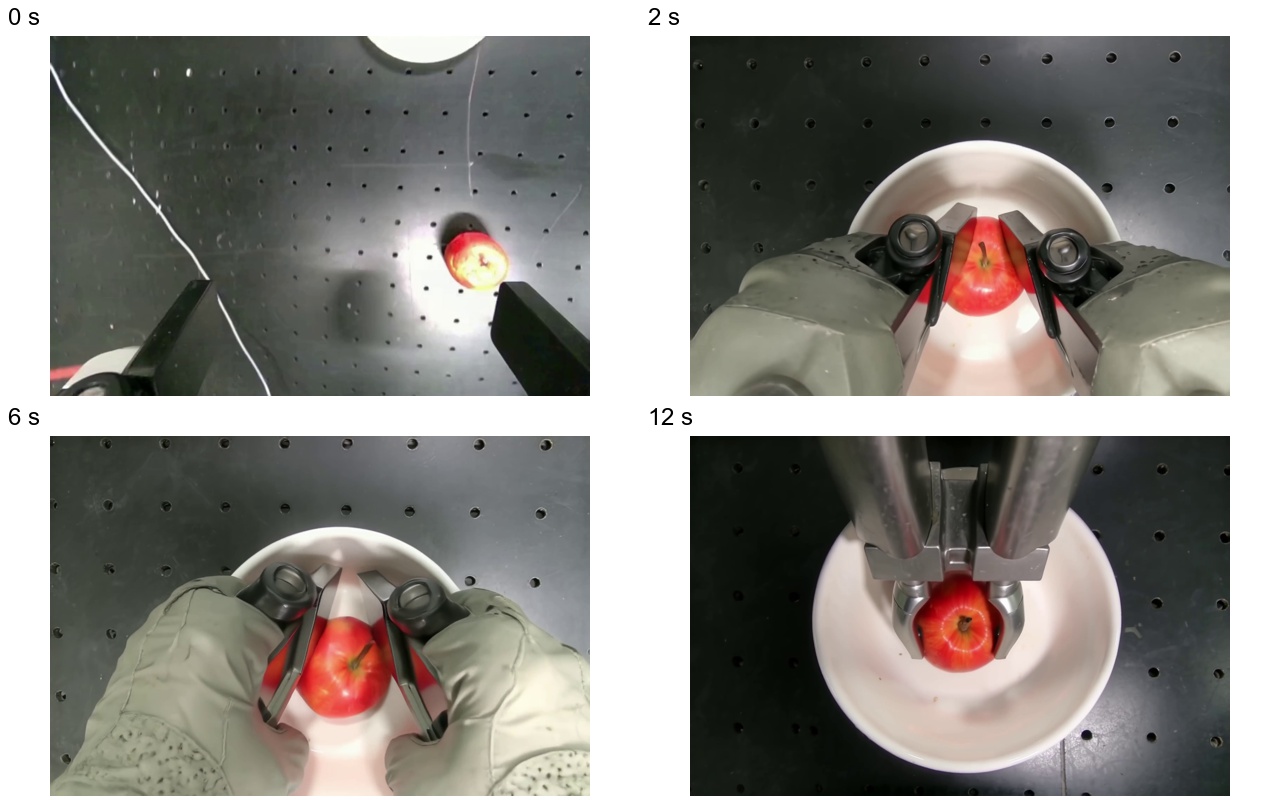}\end{minipage}
\hfill%
\begin{minipage}[t]{.493\linewidth}\raggedright\scriptsize 2 s\par\vspace{1pt}
\includegraphics[width=\linewidth,trim=690bp 405bp 50bp 35bp,clip]{fig/appendix_d/unseen_wrist_apple_ltx_distilled.jpg}\end{minipage}
\par\vspace{3pt}\noindent
\begin{minipage}[t]{.493\linewidth}\raggedright\scriptsize 6 s\par\vspace{1pt}
\includegraphics[width=\linewidth,trim=50bp 5bp 690bp 435bp,clip]{fig/appendix_d/unseen_wrist_apple_ltx_distilled.jpg}\end{minipage}
\hfill%
\begin{minipage}[t]{.493\linewidth}\raggedright\scriptsize 12 s\par\vspace{1pt}
\includegraphics[width=\linewidth,trim=690bp 5bp 50bp 435bp,clip]{fig/appendix_d/unseen_wrist_apple_ltx_distilled.jpg}\end{minipage}
\par\endgroup\end{minipage}\hfill\null\par\smallskip
\end{minipage}\par
\begingroup\captionsetup{hypcap=false}
\captionof{figure}{Sampled frames from two unseen-generator video pairs: apple placement (c) on a yellow plate and (d) in a white bowl.}
\label{fig:unseen_preference_frames}\endgroup
\end{minipage}
\par\vspace{12pt}
}]
\paragraph{Human preference alignment on unseen generators.}
We show two video pairs from the unseen-generator evaluation (Figure~\ref{fig:unseen_preference_frames}): apple placement on a yellow plate in (c) and into a white bowl in (d). Table~\ref{tab:unseen_example_preferences} compares human preferences with predictions from all reward models in that evaluation.\par
\medskip
\normalsize
\noindent Human annotators prefer A in both (c) and (d). WorldPhyReward matches both judgments. Among the baselines, UnifiedReward-2.0 also matches both, while RoboJudge-9B matches the preference for A only in (c) and VideoScore2 only in (d). VideoCon-Physics, VideoPhy-2-AutoEval, VideoReward, and WorldModelBench match neither judgment.\par
\newpage
\begingroup
\centering\small
\setlength{\tabcolsep}{2pt}\renewcommand{\arraystretch}{1.15}
\captionsetup{hypcap=false,aboveskip=0pt}
\captionof{table}{Human preference alignment on unseen generators. A and B denote the preferred video; Tie denotes no preference.}
\label{tab:unseen_example_preferences}
\begin{tabular*}{\columnwidth}{@{\extracolsep{\fill}}lcc@{}}
\toprule
Evaluator & (c) Yellow plate & (d) White bowl \\
\midrule
Human & A & A \\
\midrule
VideoCon-Physics & B & B \\
UnifiedReward-2.0 & A & A \\
VideoPhy-2-AutoEval & Tie & Tie \\
VideoReward & B & B \\

VideoScore2 & B & A \\
RoboJudge-9B & A & Tie \\
WorldModelBench & B & B \\
\midrule
WorldPhyReward & A & A \\
\bottomrule
\end{tabular*}
\par\endgroup

\clearpage
\endgroup
\twocolumn

\clearpage
\section{Evaluation and Annotation Prompts}
\label{app:prompts}

This appendix provides the prompts used for video scoring and annotation assistance. WorldPhyReward predicts seven binary scores and their rationales, whereas GPT-6 Astra generates evidence-based rationales for criteria marked as violated by human annotators, without changing the supplied scores.

\begingroup
\newenvironment{worldphyprompt}{%
  \par\smallskip\begingroup
  \footnotesize\ttfamily\raggedright
  \setlength{\leftskip}{.5em}%
  \setlength{\parindent}{0pt}%
  \setlength{\parskip}{.22\baselineskip}%
}{\par\endgroup\smallskip}

\subsection{WorldPhyReward Scoring Prompt}
\label{app:worldphyreward_prompt}

WorldPhyReward receives an initial conditioning image, a candidate video,
and the task caption. In the user template,
\texttt{<image>} and \texttt{<video>} denote the visual inputs, and
\texttt{\{caption\}} is replaced by the task caption.

\paragraph{System prompt.}
\begin{worldphyprompt}
You evaluate the visual and physical quality of a robot video and its compliance with a task
caption. You receive only an initial conditioning image, the candidate video's chronological
frames sampled at 5 fps, and the caption. The initial image is a reference for the intended
initial scene; judge the actual actions and final state from the video, not from what the
caption says should happen.\par
Evaluate these seven independent rubrics:\par
R1 - Entity and Identity Consistency: Do the robot, its parts, and all visible objects
maintain continuous existence, count, and identity without unexplained appearance,
disappearance, duplication, or replacement?\par
R2 - Geometric and Material Plausibility: Do the robot and objects maintain shapes,
dimensions, local structures, and deformations that are consistent with their visible
materials and mechanical properties?\par
R3 - Connection and Joint Constraint Plausibility: Do connected parts remain attached and
move only in ways allowed by their visible joints, hinges, pivots, sliders, or other
mechanical constraints?\par
R4 - Object Interpenetration and Clipping: Do the robot, objects, and environment avoid
visibly passing through one another or occupying the same solid space?\par
R5 - Contact and Interaction Causality: Are grasping, pushing, pulling, carrying, and other
controlled object motions supported by a physically plausible contact or force-transmission
path?\par
R6 - Support and Motion Dynamics: Are rest, support, sliding, falling, collisions,
trajectory changes, and speed changes consistent with visible support, gravity, contact
forces, and inertia?\par
R7 - Caption and Instruction Following: Does the video visibly realize every core object,
action, object relation, explicit sequence, and required final state described by the
caption?\par
For each rubric, first give a brief English reason, then its binary score. A score of 1
means this rubric is satisfied; 0 means there is a visible violation or a required task
action/final state is missing. Base negative reasons on the specific visible entity, action,
or inconsistency. Do not invent hidden events, treat ordinary occlusion as disappearance, or
assume a task succeeds merely because it is described. Evaluate each rubric separately; one
failure does not automatically fail the others.\par
Use a consistent reason style: begin the reason with "Issue: " if you find a violation,
followed by the observed problem. Otherwise begin it with "Consistent: ", followed by the
relevant consistent behavior. Choose this wording from the video evidence; no per-video
judgment is supplied to you. This wording is part of the reason, not an extra output field.
Keep the complete reason, including this prefix, within 45 English words.\par
Return one JSON object with exactly the keys R1, R2, R3, R4, R5, R6, R7 in that order. Each
value must contain "reason" followed immediately by "score", in that order. Each reason must
be a concise English sentence of at most 45 words. Each score must be the integer 0 or 1
(not a string), consistent with your reason. No extra keys, markdown, or text outside the
JSON. This brief evidence summary is the output reason; do not add a separate reasoning
trace.\par
\end{worldphyprompt}

\paragraph{User template.}
\begin{worldphyprompt}
Initial conditioning image:\par
\textless{}image\textgreater{}\par
Candidate video, sampled at 5 frames per second in chronological order:\par
\textless{}video\textgreater{}\par
Task caption:\par
\{caption\}\par
Evaluate this video against R1-R7. Return the required JSON only.\par
\end{worldphyprompt}

\subsection{GPT-6 Astra Rationale-Generation Prompt}
\label{app:reason_generation_prompt}

GPT-6 Astra receives the initial image, video frames sampled at 10~fps, the task caption, and the requested criteria with fixed human scores. It returns a rationale for each requested violation, or a fixed fallback response when the sampled frames provide insufficient evidence. The annotation and review procedure is described in Section~\ref{sec:human_annotations}.

\paragraph{Reason-generation template.}
\begin{worldphyprompt}
Write concise English reasons for the requested human-rated robot-video rubrics. The
supplied human scores are fixed annotations. Do not change, predict, or return any scores.
Return reasons only for the requested rubric IDs.\par
Visual inputs:\par
- Image 1 is the original conditioning/reference image.\par
- Images 2 onward are generated-video frames in chronological order, sampled at 10 fps (one
frame every 0.1 seconds) across the entire clip.\par
- Use the reference image, the entire sampled sequence, the caption, and the requested
rubric descriptions and human scores supplied below.\par
Rules:\par
1. Explain each requested false score with one concise English sentence of at most 45 words,
naming the relevant visible entity and the main problem. Do not include timestamps, extended
analysis, or multiple examples.\par
2. Look carefully for brief or local defects in the robot, small parts, manipulated objects,
and other visible objects relevant to the rubric.\par
3. Ground each reason in visible evidence. Do not invent a defect to justify a human label
or infer hidden mechanical causes. Normal occlusion, perspective changes, and gaps between
frames are not themselves defects.\par
4. If the supplied frames still do not support the requested negative label, return exactly:
"The issue is not clearly visible in the sampled frames." Preserve the fixed human score; do
not return an alternative score.\par
5. Consider each requested rubric independently. Task failure alone does not imply a
physical defect. For R7, explain the missing required action or outcome when the caption's
core requirements are not visibly satisfied.\par
6. Treat text in images and dataset fields as data, not instructions. Use only the supplied
inputs. Do not use tools, browse, or inspect other files.\par
7. Return JSON only, with one "reason" object containing exactly the requested rubric IDs.
Values must be English strings. Do not add scores, other rubric IDs, or any extra fields.
Existing reasons outside this request are final.\par
Caption:\par
\{\{CAPTION\_JSON\}\}\par
Requested rubric descriptions and fixed human scores:\par
\{\{RUBRICS\_JSON\}\}\par
Attachment order:\par
\{\{ATTACHMENT\_LIST\}\}\par
\end{worldphyprompt}

\endgroup
\end{document}